\documentclass{article}

\usepackage{pratik_icml}
\usepackage[accepted]{icml2023}

\def \Pt {P_t}
\def \Pout {P_o}

\DeclareMathOperator*{\constant}{\text{const.}}

\graphicspath{{fig/}{figures/}}

\makeatletter
\DeclareMathSizes{\@xpt}{9}{7}{5}
\makeatother

\icmltitlerunning{\hfill The Value of Out-of-Distribution Data\hfill\thepage}

\begin{document}


\twocolumn[
\icmltitle{The Value of Out-of-Distribution Data}
\icmlsetsymbol{equal}{*}
\icmlsetsymbol{equal2}{\dag}

\begin{icmlauthorlist}
\icmlauthor{Ashwin De Silva}{jhu,equal}
\icmlauthor{Rahul Ramesh}{upenn,equal}
\icmlauthor{Carey E. Priebe}{jhu}
\icmlauthor{Pratik Chaudhari}{upenn,equal2}
\icmlauthor{Joshua T. Vogelstein}{jhu,equal2}
\end{icmlauthorlist}

\icmlaffiliation{upenn}{University of Pennsylvania}
\icmlaffiliation{jhu}{Johns Hopkins University}

\icmlcorrespondingauthor{Ashwin De Silva}{ldesilv2@jhu.edu}
\icmlcorrespondingauthor{Rahul Ramesh}{rahulram@seas.upenn.edu}


\vskip 0.3in
]



\printAffiliationsAndNotice{\icmlEqualContribution} 


\begin{abstract}

Generalization error always improves with more in-distribution data. However, it is an open question what happens as we add out-of-distribution (OOD) data.  Intuitively, if the OOD data is quite different, it seems more data would harm generalization error, though if the OOD data are sufficiently similar, much empirical evidence suggests that OOD data can actually improve generalization error. 
We show a counter-intuitive phenomenon: \textbf{the generalization error of a task can be a non-monotonic function of the amount of OOD data}. Specifically, we show that generalization error can improve with small amounts of OOD data, and then get worse with larger amounts compared to no OOD data. 
In other words, there is value in training on small amounts of OOD data. 
We analytically demonstrate these results via  Fisher's Linear Discriminant on synthetic datasets, and empirically demonstrate them via deep networks on computer vision benchmarks such as MNIST, CIFAR-10, CINIC-10, PACS and DomainNet. 
In the idealistic setting where we know which samples are OOD, we show that these non-monotonic trends can be exploited using an appropriately weighted objective of the target and OOD empirical risk. While its practical utility is limited, this does suggest that \emph{if} we can detect OOD samples, then there may be ways to benefit from them. When we do not know which samples are OOD, we show how a number of go-to strategies such as data-augmentation, hyper-parameter optimization and pre-training are not enough to ensure that the target generalization error does not deteriorate with the number of OOD samples in the dataset.

\end{abstract}
\section{Introduction}


Real data is often heterogeneous and more often than not, suffers from distribution shifts. We can model this heterogeneity as samples drawn from a mixture of a target distributrion and from ``out-of-distribution'' (OOD). For a model trained on such data, we expect one of the following outcomes: (i) if the OOD data is similar to the target data, then more OOD samples will help us generalize to the target distribution; (ii) if the OOD data is dissimilar to the target data, then more samples are detrimental. In other words, we expect the target generalization error to be \emph{monotonic} in the number of OOD samples; this is indeed the rationale behind classical works such as that of~\citet{ben2010theory} recommending against having OOD samples in the training data.

We show that a third counter-intuitive possibility occurs: OOD data from the \textbf{same} distribution can both improve or deteriorate the target generalization depending on the number of OOD samples. \textbf{Generalization error (note: error, not the gap) on the target task is non-monotonic in the number of OOD samples.} Across  numerous examples, we find that there exists a threshold below which OOD samples improve generalization error on the target task but if the number of OOD samples is beyond this threshold, then the generalization error deteriorates. To our knowledge, this phenomenon has not been predicted or demonstrated by any other theoretical or empirical result in the literature.

We first demonstrate the non-monotonic behavior through a simple but theoretically tractable problem using Fisher's Linear Discriminant (FLD). In~\cref{ss:upper_bound_violation}, for the same problem, we compare the actual expected target generalization error with the theoretical upper bound developed by~\citep{ben2010theory} to show that this phenomenon is not captured by existing theory.
\textbf{We also present empirical evidence for the presence of non-monotonic trends in target generalization error, on tasks and experimental settings constructed from the MNIST, CIFAR-10, PACS and DomainNet datasets}. Our code is available at \href{https://github.com/neurodata/value-of-ood-data}{https://github.com/neurodata/value-of-ood-data}.

\subsection{Outlook}

Consider the idealistic setting where we know which samples in the dataset are OOD. A trivial solution could be to remove the OOD samples from the training set. But the fact that the generalization error is non-monotonic also suggests a better solution. We show on a number of benchmark tasks that by using an appropriately weighted objective between the target and OOD samples, we can ensure that the generalization error on the target task decreases monotonically with the number of OOD samples. This is merely a proof-of-concept for this idealistic setting. But it does suggest that \emph{if} one could detect the OOD samples, then there are not only ways to safeguard against them but there are also ways to benefit from them.

Of course, we do not know which samples are OOD in real datasets. When datasets are curated incrementally, the fraction of OOD samples can also change with time, and the implicit benefit of these OOD data may become a drawback later. When we do not know which samples are OOD, we show how a number of go-to strategies such as data-augmentation, hyper-parameter optimization and pre-training the network are not enough to ensure that the generalization error on the target does not deteriorate with the number of OOD samples. 

Our results indicate that non-monotonic trends in generalization error are a significant concern, especially when the presence of OOD samples in the dataset goes undetected. The main contribution of this paper is to highlight the importance of this phenomenon. We leave the development of a practical solution for future work.

\section{Generalization error is non-monotonic in the number of OOD samples}
\label{s:task_agnostic_setting}

We define a distribution $P$ as a joint distribution over the input domain $X$ and the output domain $Y$. We model the heterogeneity in the dataset as two distributions: $n$ samples drawn from a target distribution $\Pt$ and $m$ samples drawn from out-of-distribution (OOD) $\Pout$. We would like to minimize the generalization error \( e_t(h) =  \E_{(x, y) \sim \Pt} \sbr{h(x) \neq y} \) on the target distribution. Suppose we assume that all the data comes from a single target distribution because we are unaware of the presence of OOD samples in the dataset. Therefore, we may find a hypothesis that minimizes the empirical loss
\beq{
    \hat e(h) =  \f{1}{n+ m}\sum_{i=1}^{n+m} \ell \rbr{h(x_i), y_i},
    \label{eq:task_agnostic_erm}
}
using the dataset $\cbr{(x_i, y_i)}_{i=1}^{n+m}$; here $\ell$ measures the mismatch between the prediction $h(x_i)$ and label $y_i$. If $\Pt = \Pout$, then $e_t(h) - \hat e(h) = \OO((n+m)^{-1/2})$~\citep{smola1998learning}. But if $\Pt$ is far enough from $\Pout$ in certain ways, then we expect that the error on $\Pt$ of a hypothesis obtained by minimizing the average empirical loss will be suboptimal, especially when the number of OOD samples $m \gg n$.

\subsection{An example using Fisher's Linear Discriminant}
\label{s:fld_example}

\begin{figure}[!b]
\centering
\includegraphics[width=\linewidth]{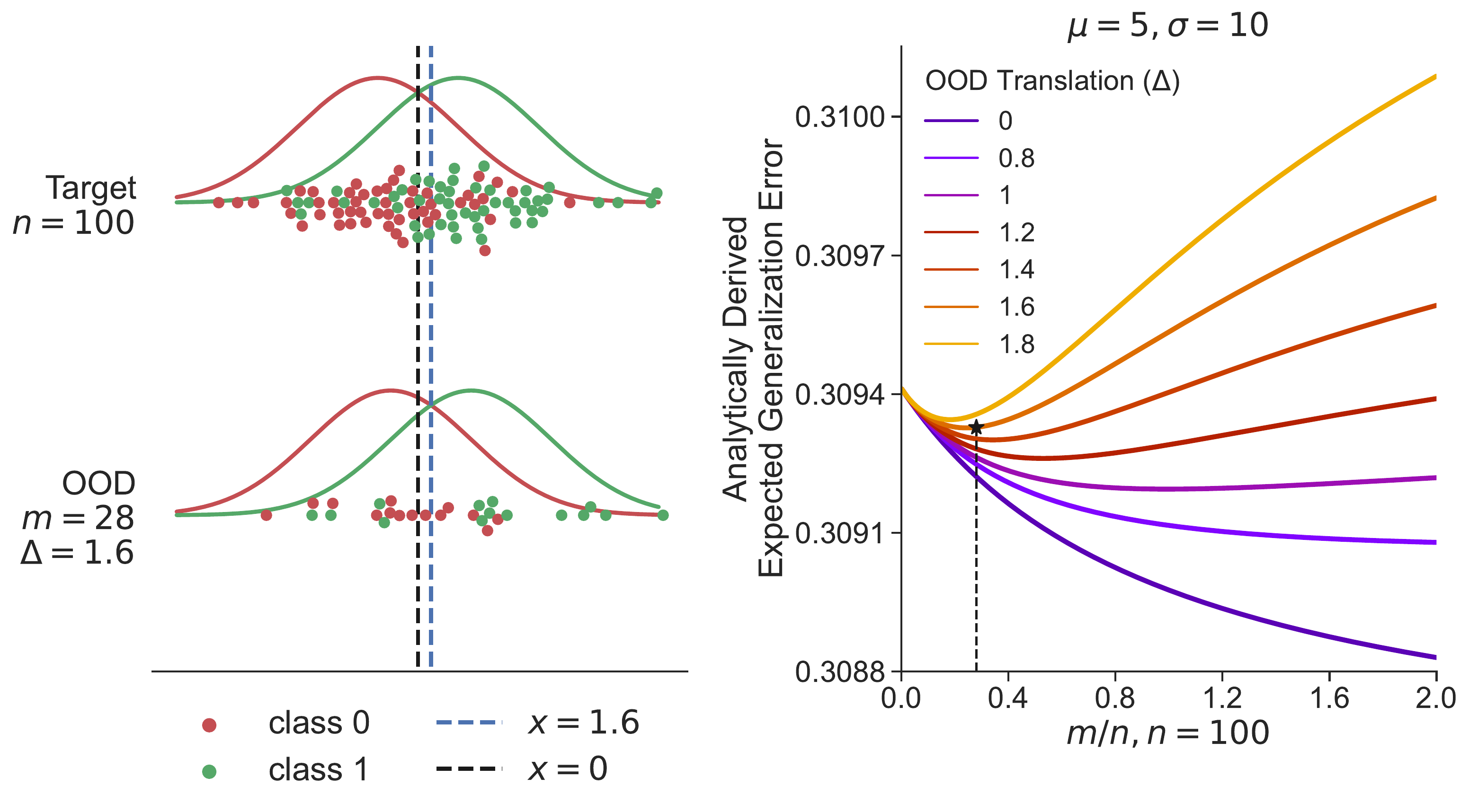}
\caption{
\textbf{Left:} A schematic of the Gaussian mixture model corresponding to the target (top) and OOD samples (bottom). The OOD sample size ($m=28$) at which the target generalization error is minimized at $\Delta = 1.6$ is indicated at the top.
\textbf{Right:} For $n =100$, we plot the generalization error of FLD on the target distribution as a function of the ratio of OOD and target samples $m/n$, for different types of OOD samples corresponding to different values of $\D$. This plot uses the analytical expression for the generalization error in~\cref{eqn:fld_target_gen_err}; see~\cref{app:fld_simualtions} for a numerical simulation study. For small values of $\D$, when the two distributions are similar to each other, the generalization error $e_t(h)$ decreases monotonically. However, beyond a certain value of $\D$, the generalization error is non-monotonic in the number of OOD samples. The optimal value of $m/n$ which leads to the best generalization error is a function of the relatedness between the two distributions, as governed by $\D$ in this example. This non-monotonic behavior can be explained in terms of a bias-variance tradeoff with respect to the target distribution: a large number of OOD samples reduces the variance but also results in a bias with respect to the optimal hypothesis of the target.
}
\label{fig:gauss_tasks}
\end{figure}

Consider a binary classification problem with one-dimensional inputs in~\cref{fig:gauss_tasks}. Target samples are drawn from a Gaussian mixture model (with means $\cbr{-\mu, \mu}$ for the two classes) and OOD samples are drawn from a Gaussian mixture with means $\cbr{-\mu +\D, \mu+\D}$; see~\cref{sec:app:synthetic} for details. Fisher's linear discriminant (FLD) is a linear classifier for binary classification problems and it computes $\hat h(x) = 1$ if $\omega^\top x > c$ and $\hat h(x) = 0$ otherwise; here $\omega$ is a projection vector which acts as a feature extractor and $c$ is a threshold that performs one-dimensional discrimination between the two classes. FLD is optimal when  the class conditional density of each class is a multivariate Gaussian distribution with the same covariance structure. We provide a detailed account of FLD in~\cref{app:fld_derivation}.

Suppose we fit an FLD on a dataset which comprises of $n$ target samples and $m$ OOD samples. Also, suppose we do not know which samples are OOD and believe that all the samples in the dataset come from a single target distribution. For univariate data with equal class priors, the FLD decision rule reduces to,
\begin{equation}
    \hat h(x) =
    \begin{cases}
        1, & x > \frac{\hat \mu_0 + \hat \mu_1}{2} \\
        0, & \text{otherwise}.
    \end{cases} \notag
\end{equation}
Define the decision threshold to be $\hat c = (\hat \mu_0 + \hat \mu_1)/2$. We can calculate (\cref{app:fld_derivation,app:fld_err}) an analytical expression for the generalization error of FLD on the target distribution:
\begin{align}
    \scalemath{0.74}{
    e_t(\hat{h}) = \frac{1}{2}\bigg[ \Phi \bigg( \frac{m \Delta - (n + m)\mu}{\sqrt{(n+m)(n+m+1)}} \bigg) +
    \Phi \bigg( \frac{-m \Delta - (n + m)\mu}{\sqrt{(n+m)(n+m+1)}} \bigg)\bigg];}
    \label{eqn:fld_target_gen_err}
\end{align}
here $\Phi$ is the CDF of the standard normal distribution.

\begin{figure}[htb]
\centering
\includegraphics[width=0.45\linewidth]{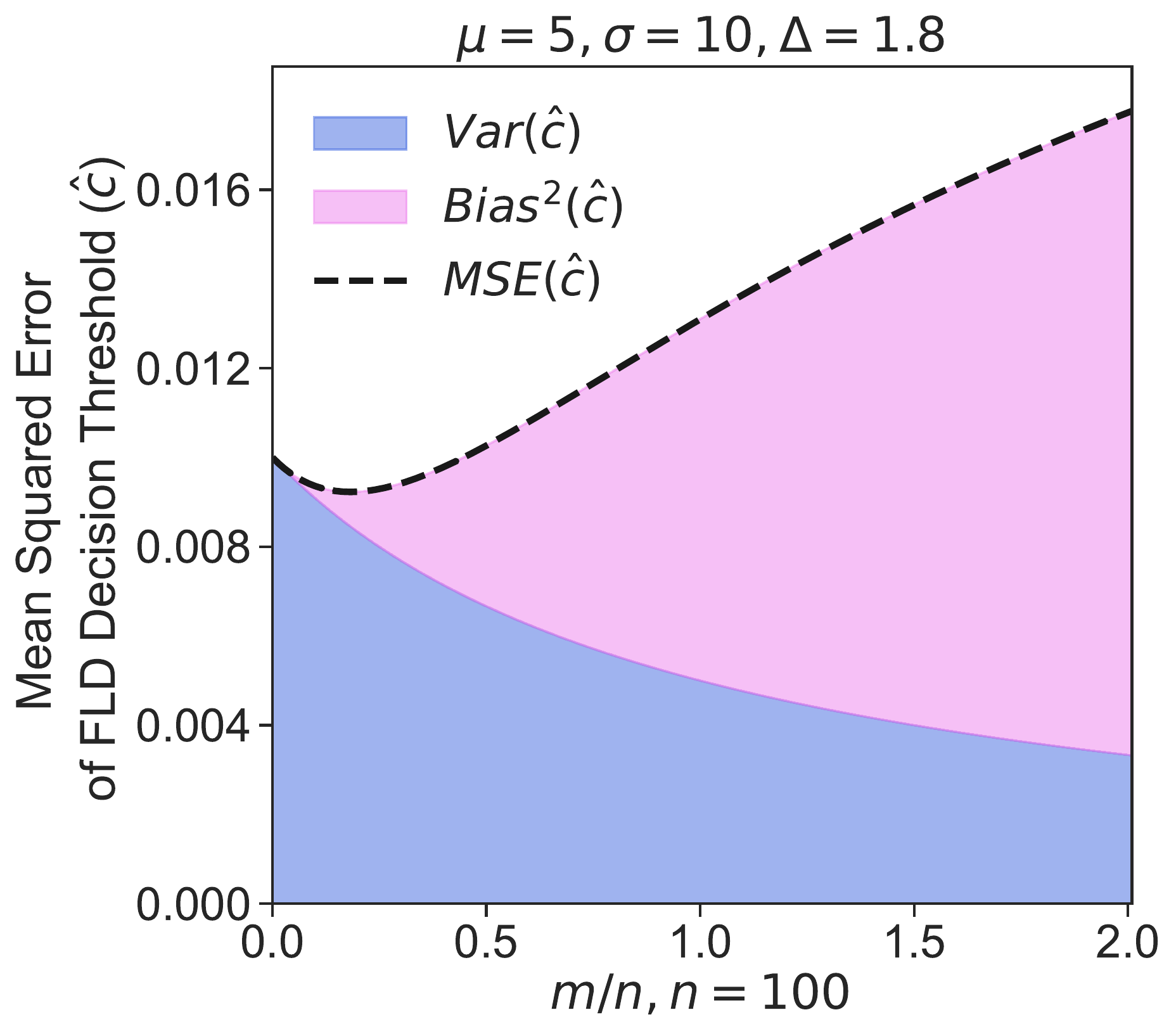}
\caption{\textbf{Mean squared error (MSE) (Y-axis) of the decision threshold $\hat c$ of FLD} (see~\cref{app:fld_err}), for the same setup as that of~\cref{fig:gauss_tasks}, plotted against the ratio of the OOD and target samples $m/n$ (X-axis) for $\Delta = 1.8$. Squared bias and variance of the MSE are in violet and blue, respectively. This illustration clearly demonstrates the intuition behind non-monotonic target error: the MSE drops initially because of the smaller variance due to the OOD samples. With more OOD samples, MSE increases due to the increasing bias. Non-monotonic trend in MSE of $\hat c$ translates to a similar trend in the target generalization error (0-1 loss). }
\label{fig:fld_bias_var}
\end{figure}

\begin{figure}[htb]
\centering
\vspace{-2em}
\includegraphics[width=0.45\linewidth]{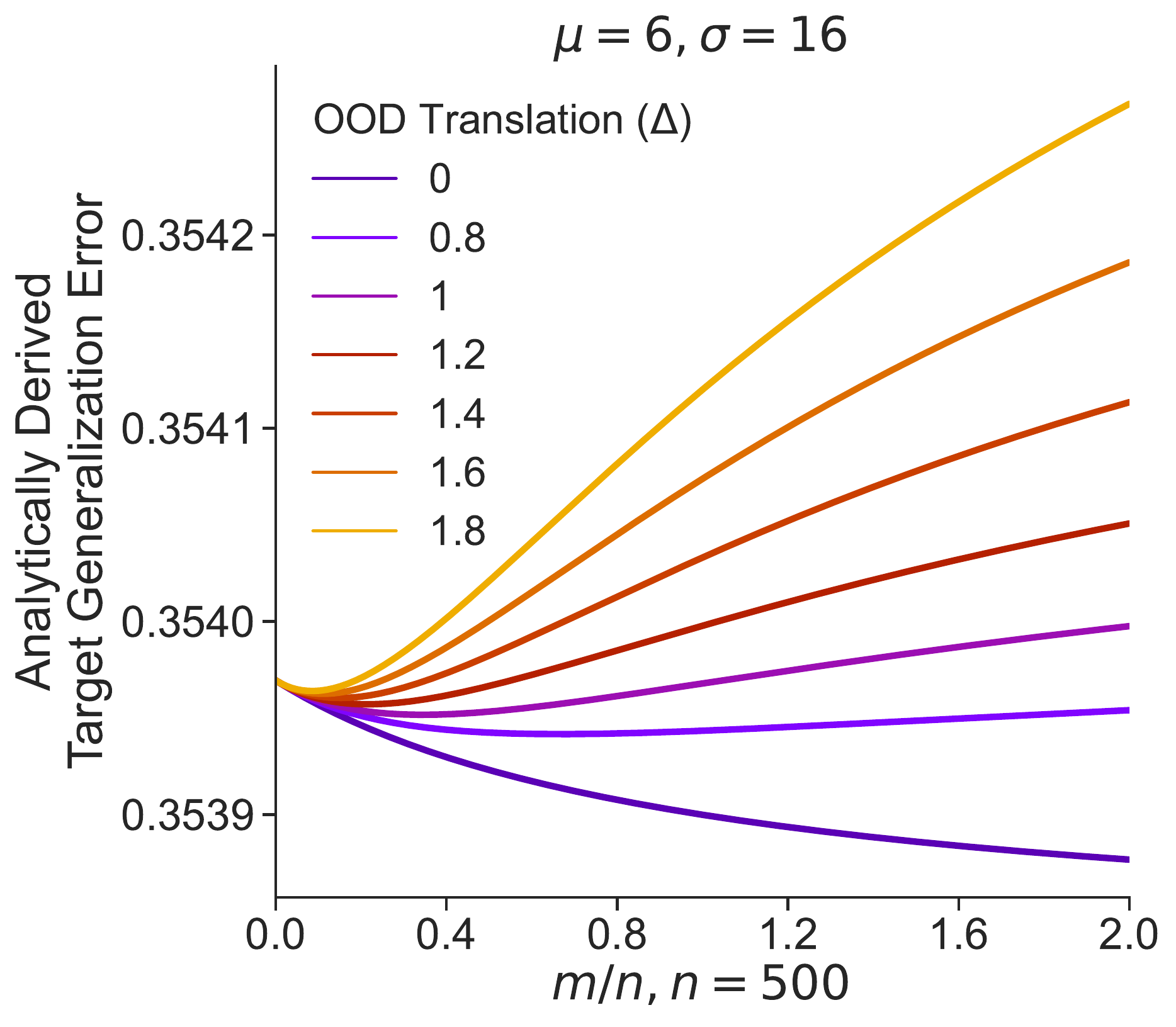}
\caption{We can control the Bayes optimal error by adjusting $\mu, \sigma$ of the Gaussian mixture model in~\cref{s:fld_example}. As discussed in~\cref{rem:intuitive}, when the Bayes optimal error is large for $(\mu=6,\sigma=16)$, we can observe non-monotonic trends even for a large number of target samples ($n=500)$. This suggests that non-monotonic trends in generalization are not limited to small sample sizes.}
\vspace{-0.5em}
\label{fig:large_n}
\end{figure}

\cref{fig:gauss_tasks} (right) shows how the generalization error $e_t(\hat{h})$ decreases up to some threshold of the ratio between the number of OOD samples and the number of target samples $m/n$ and then increases beyond that. This threshold is different for different values of $\D$ as one can see in~\cref{eqn:fld_target_gen_err} and~\cref{fig:gauss_tasks} (right). This behavior is surprising because one would \emph{a priori} expect the generalization error to be monotonic in the number of OOD samples. The fact that a non-monotonic trend is observed even for a one dimensional Gaussian mixture model suggests that this may be a general phenomenon. We can capture this discussion as a theorem; the FLD example above is the proof.
\begin{theorem}
There exist target and OOD distributions, $\Pt$ and $\Pout$ respectively, such that the generalization error on the target distribution of the hypothesis that minimizes the empirical loss in~\cref{eq:task_agnostic_erm}, is non-monotonic in the number of OOD samples. In particular, there exist distributions $\Pt$ and $\Pout$ such that the generalization error decreases with few OOD samples and increases with even more OOD samples, compared to no OOD samples.
\end{theorem}

\begin{remark}[An intuitive explanation of non-monotonic trends in generalization error]
\label{rem:intuitive}
Suppose that a learning algorithm achieves Bayes optimal error on the target distribution with high probability when the target sample size $n$ exceeds $N$.
We argue that a non-monotonic trend in generalization error is likely to occur when $n < N$, i.e., when target generalization error is higher than the Bayes optimal error. In this case, if we add OOD samples whose empirical distribution is sufficiently close to that of the target distribution, then this would improve generalization by reducing the variance of the learned hypothesis. But as the OOD sample size increases, the difference between the two distributions becomes apparent and this leads to a bias in the choice of the hypothesis. \cref{fig:fld_bias_var} illustrates this phenomenon with regards to our FLD example in~\cref{fig:gauss_tasks}, by plotting the mean squared error of the decision threshold $\hat c$ and its constituent bias and variance components. Roughly speaking, we may understand the non-monotonic trend in generalization as a phenomenon that arises due to the finite number of OOD samples ($m/n$ in the example above). The distance between the distribution of the OOD samples and the distribution of the target samples ($\D$ in the example) determines the threshold beyond which the error is monotonic. Current tools in learning theory~\citep{smola1998learning} are fundamentally about understanding generalization when the number of samples is asymptotically large---whether they be from the target or OOD. In future work, we hope to formally characterize this non-monotonic trend in generalization error by building new learning-theoretic tools.
\end{remark}

Even if the non-monotonic trend occurs for relatively small values of target and OOD samples $n$ and $m$ respectively in~\cref{fig:gauss_tasks}, this need not always be the case. \textbf{If the number of samples $N$ required to reach Bayes optimal error in the above remark is large, then a non-monotonic trend can occur even for large target sample size $n$ (see~\cref{fig:large_n})}.

\subsection{Non-monotonic trends for neural networks and machine learning benchmark datasets}
\label{s:non-monotonic_nn}

\begin{figure*}[!t]
\centering
\includegraphics[width=0.65\linewidth]{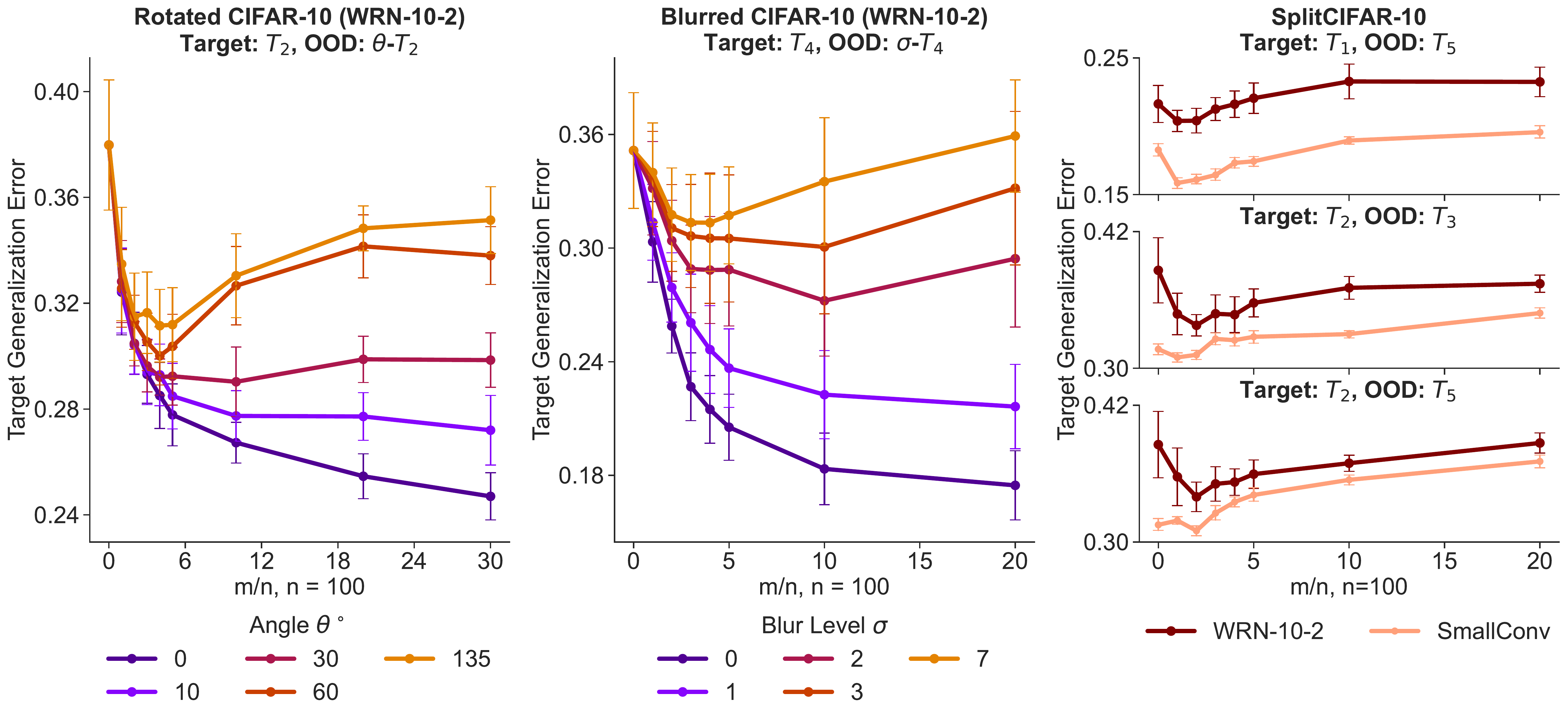}
\caption{\textbf{Left:} Sub-task $T_2$ (Bird vs. Cat) from Split-CIFAR10 is the target data and images of these classes rotated by different angles $\theta^\circ$ are the OOD data. WRN-10-2 architecture was used to train the model. We see non-monotonic curves for larger values of $\theta^\circ$. For $60^\circ$ and $135^\circ$ in particular, the generalization error at $m/n=20$ is worse than the generalization error with a fewer OOD samples, i.e. OOD samples actively hurt generalization. See~\cref{fig:selected_rotated_and_dual_tasks} (left) for a similar experiment with SmallConv.\\[0.1em]
\textbf{Middle:} The Split-CIFAR10 binary sub-task $T_4$ (Frog vs. Horse) is the target distribution and images with different levels of Gaussian blur are the OOD samples. WRN-10-2 architecture was used to train the model. Non-monotonic curves are observed for larger levels of blur, while for smaller levels of blur, we notice that adding more OOD data improves the generalization on the target distribution.\\[0.1em]
\textbf{Right:} Generalization error of two separate networks, WRN-10-2 and SmallConv, on the target distribution is plotted against the number of OOD samples for 3 different target-OOD pairs from Split-CIFAR10. All the 3 pairs exhibit non-monotonic target generalization trends across both network models. See~\cref{app:expt_details,app:form-tar-ood} for experimental details and~\cref{app:splitcifar10_task_matrix} for experiments on more target-OOD pairs (\cref{fig:cifar1_task_matrix,fig:cifar2_task_matrix}) and multiple target sample sizes (\cref{fig:multiple_n_exp}).
Error bars indicate 95\% confidence intervals (10 runs).
}
\label{fig:non-monotonicity-fig-1}
\end{figure*}

We experiment with several popular datasets including MNIST, CIFAR-10, PACS, and DomainNet and 3 different network architectures: (a) a small convolutional network with 0.12M parameters (denoted by \emph{SmallConv}), (b) a wide residual network~\citep{zagoruyko2016wide} of depth 10 and widening factor 2 (WRN-10-2), and (c) a larger wide residual network of depth 16 and widening factor 4 (WRN-16-4). See~\cref{sec:app:arch} for more details.

\paragraph{A non-monotonic trend in generalization error can occur due to geometric and semantic nuisances.}
Such nuisances are very common even in curated datasets~\citep{van2019towards}. We constructed 5 binary classification sub-tasks (denoted by $T_i$ for  $i=1,\ldots,5$) from CIFAR-10 to study this aspect (see~\cref{sec:app:image}). We consider a CIFAR-10 sub-task $T_2$ (Bird vs.\@ Cat) as the target and introduce rotated images by a fixed angle between $0^\circ$-$135^\circ$) as OOD samples. \cref{fig:non-monotonicity-fig-1} (left) shows that the generalization error decreases monotonically for small rotations but it is non-monotonic for larger angles. Next, we considered the sub-task $T_4$ (Frog vs.\@ Horse) as the target distribution and generate OOD samples by adding Gaussian blur of varying levels to images from the same distribution. In~\cref{fig:non-monotonicity-fig-1} (middle), the generalization error on the target is a monotonically decreasing function of the number of OOD samples for low blur but it increases non-monotonically for high blur.

\vspace{-0.5em}
\paragraph{Non-monotonic trends can occur when OOD samples are drawn from a different distribution}
Large datasets can contain categories whose appearance evolves in time (e.g., a typical laptop in 2022 looks very different from that of 1992), or categories can have semantic intra-class nuisances (e.g., chairs of different shapes). We use 5 CIFAR-10 sub-tasks to study how such differences can lead to non-monotonic trends (see~\cref{sec:app:image}). Each sub-task is a binary classification problem with two consecutive classes: Airplane vs.\@ Automobile, Bird vs.\@ Cat, etc. We consider $(T_i, T_j)$ as the (target, OOD) pair and evaluated the trend in generalization error for all 20 distinct pairs of distributions. \cref{fig:non-monotonicity-fig-1} (right) illustrates non-monotonic trends for 3 such pairs; see~\cref{sec:app:neural_net} for more details.

\begin{figure*}[htpb]
\centering
\includegraphics[width=0.7\textwidth]{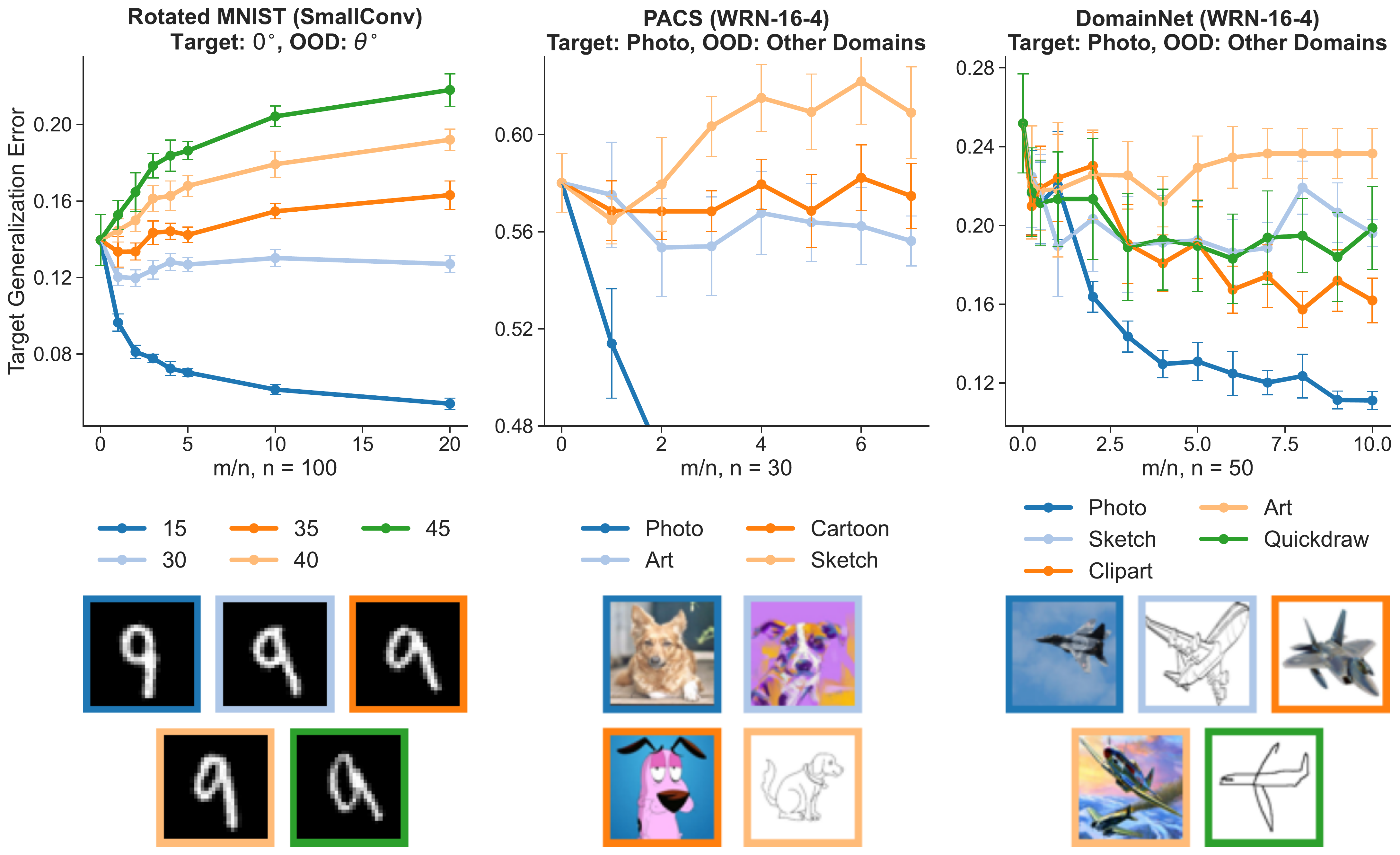}
\caption{
\textbf{Non-monotonic trends in target generalization error on three DomainBed benchmarks.}
Left: Rotated MNIST (10 classes, 10 target samples/class, \emph{SmallConv}),
Middle: PACS (3 classes \{dog, elephant, horse\}, 10 target samples/class, WRN-16-4), and
Right: DomainNet (2 classes \{bird, plane\}, 25 target samples/class, WRN-16-4).
Error bars indicate 95\% confidence intervals (10 runs). Also see \cref{fig:domainnet-40} for results from a 40-way classification task from DomainNet. 
}
\label{tab:non-monotonicity-table}
\end{figure*}

\paragraph{Non-monotonic trends also occur for benchmark domain generalization datasets}
We further investigated three widely used benchmarks in the domain generalization literature. First, we consider the Rotated MNIST benchmark from DomainBed~\citep{gulrajani2020search}. We define the 10-way classification of un-rotated MNIST images as the target distribution and $\theta$-rotated MNIST images as the OOD samples. Similar to the previous rotated CIFAR-10 experiment, we observe non-monotonic trends in target generalization for larger angles $\theta$. Next, we consider the PACS benchmark from DomainBed which contains 4 distinct environments: photo, art, cartoon, and sketch. A 3-way classification task involving photos (real images) is defined as the target distribution, and we let the corresponding data from other environments be the OOD samples. Interestingly, we observe that when OOD samples consist of sketched images, then the generalization error on the real images exhibits a non-monotonic trend. We also observe similar trends in DomainNet, a benchmark that resembles PACS; see~\cref{tab:non-monotonicity-table}.

\begin{figure}[!tb]
\centering
\includegraphics[width=0.45\linewidth]{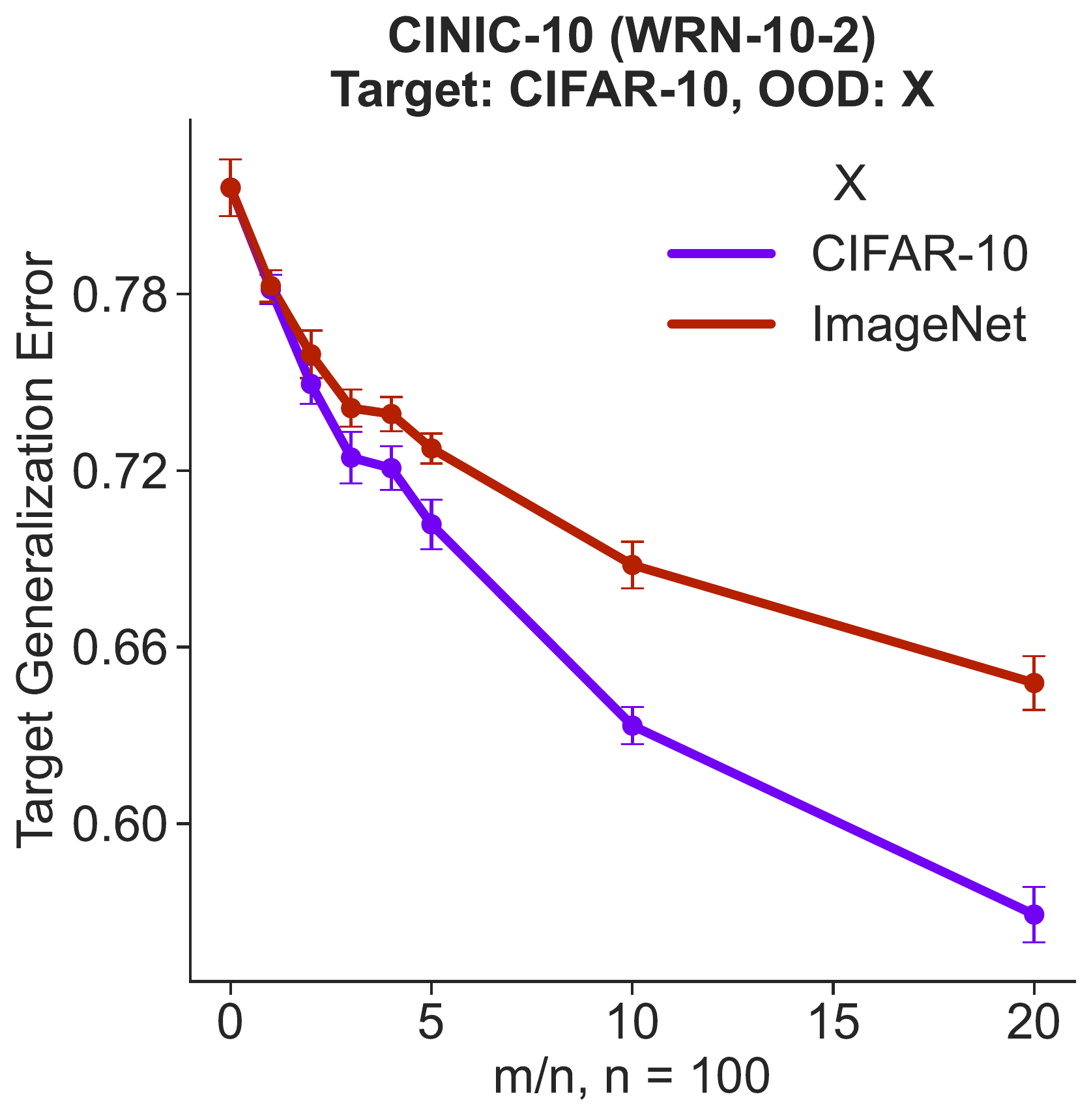}
\caption{Target task is CIFAR-10 and OOD samples are from ImageNet. Although there is a distribution shift that causes the red curve to be higher error than the purple one, there is no non-monotonic trend in the generalization on CIFAR-10 due to OOD samples from ImageNet. Error bars indicate 95\% confidence intervals (10 runs).\\}
\vspace{-1.5em}
\label{fig:cinic10_curve}
\end{figure}

\paragraph{Generalization error is not always non-monotonic even when there is distribution shift}
We considered CINIC-10~\citep{darlow2018cinic}, a dataset which was created by combining CIFAR-10 with images selected and down-sampled from ImageNet. We train a network on a subset of CINIC-10 that comprises of both CIFAR-10 and ImageNet images. The target task is CIFAR-10 itself, so images from ImageNet in CINIC-10 act as OOD samples. \cref{fig:cinic10_curve} demonstrates that having more ImageNet samples in the training data improves the generalization (monotonic decrease) on the target distribution, but at a slower rate than the instance where the training data is purely comprised of target data. This phenomenon is also demonstrated in~\cref{fig:gauss_tasks}: for sufficiently small shifts, the target generalization error decreases as the number of OOD samples increases.

\vspace{-0.5em}

\paragraph{Effect of pre-training, data-augmentation and hyperparameter optimization}
When we do not know which samples are OOD, we do not have a lot of options to mitigate the deterioration due to the OOD samples. We could use data augmentations, hyper-parameter optimization, or pre-training followed by fine-tuning. The second option is difficult to implement for a real problem because the validation data that will be used for hyper-parameter optimization will itself have to be drawn from the curated dataset. 

\begin{figure}[ht!]
\centering
\includegraphics[width=0.90\linewidth]{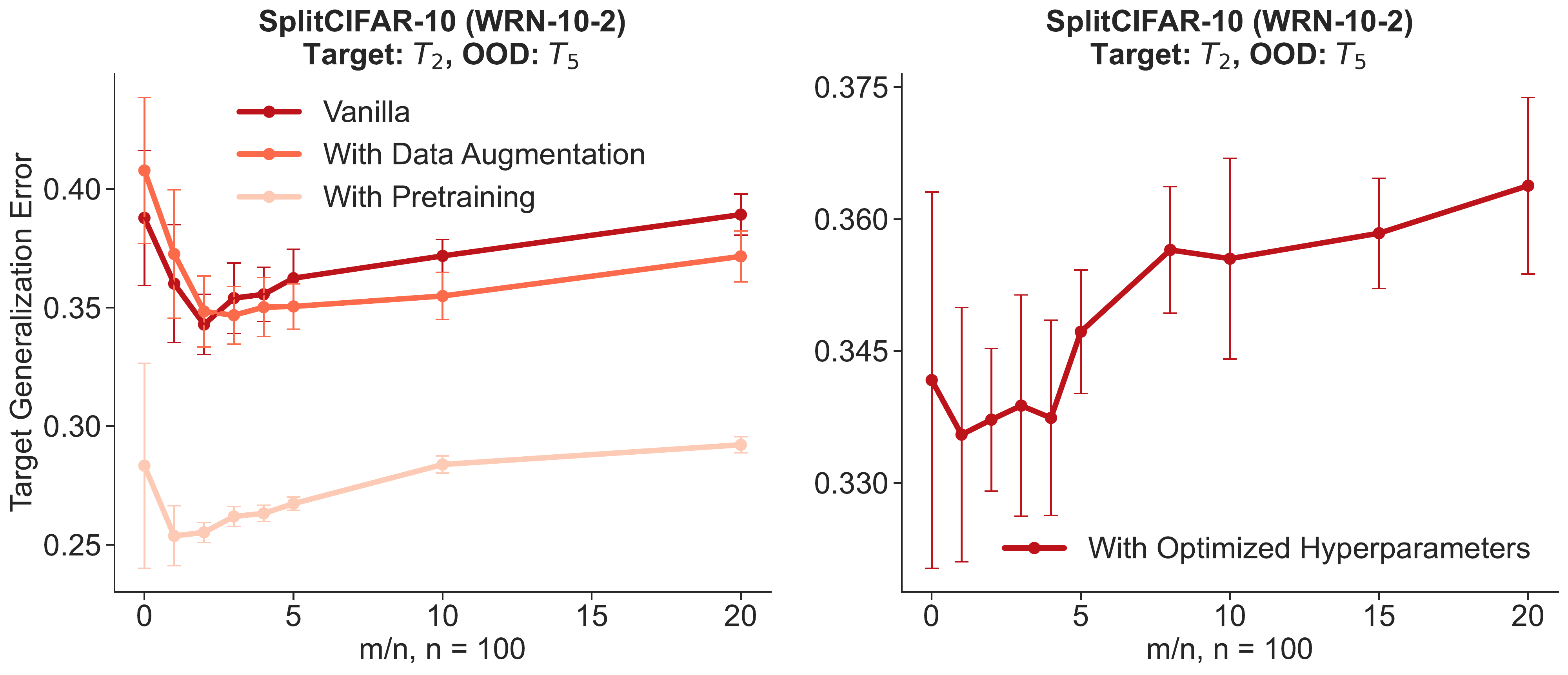}
\caption{
\textbf{Left:} For the CIFAR-10 sub-task $T_2$ (Bird vs Cat) as target and $T_5$ (Ship vs Truck) as OOD, we train a WRN-10-2 network with class-balanced datasets with fixed number of target samples ($n=100$) and different number ($m$) of OOD samples, under the following settings: \begin{enumerate*}[{(1)}] \item Vanilla, i.e., without any data-augmentation or pre-training (darkest red), \item Data augmentation by padding, random cropping and random left/right flips (medium red), and \item Pre-training followed by fine-tuning (lightest red). \end{enumerate*} We pre-train the network on $14000$ class-balanced ImageNet images from CINIC-10 (see~\cref{sec:app:image}) belonging to Bird and Cat classes which correspond to our hypothetical target distribution. Pre-training is performed for $100$ epochs with a learning rate of $0.01$. Next, we employ a two-step strategy of linear probing (first $50$ epochs) and full-fine tuning (last $50$ epochs) inspired by~\cite{kumar2022fine} at a reduced learning rate of $0.001$. Note that this fine-tuning is performed on the combined dataset of $n$ target and $m$ OOD samples. Even though data augmentation and pre-training followed by fine-tuning reduce the overall error, the generalization error still deteriorates as the fraction of OOD sample in the dataset increases. \textbf{Right:} For each value of $m$, we perform hyper-parameter tuning using Ray~\citep{liaw2018tune} over a validation set that has \emph{only} target samples, and record the target generalization error of the model using the best set of hyper-parameters. We still observe deterioration of the target generalization error as the OOD samples increase. \textbf{Note that such hyper-parameter tuning cannot be implemented in reality because we may not know the identity of the target and OOD samples}. So the fact that the non-monotonic trend persists in the hypothetical instance where we \emph{know} the sample identities guarantees that it will occur in practice as well. Error bars indicate 95\% confidence intervals over 10 experiments.\\
}
\vspace{-15pt}
\label{fig:effect-of-techniques}
\end{figure}

To evaluate whether these three techniques work, we used the CIFAR-10 sub-task $T_2$ (Bird vs.\@ Cat) as the target distribution and $T_5$ (Ship vs.\@ Truck) as the distribution of the OOD data and trained a WRN-10-2 network under various settings. The results are reported in~\cref{fig:effect-of-techniques}; we find that these techniques do not mitigate the deterioration of target generalization error as the number of OOD samples in the dataset increases.

\begin{figure}[!h]
\centering
\includegraphics[width=\linewidth]{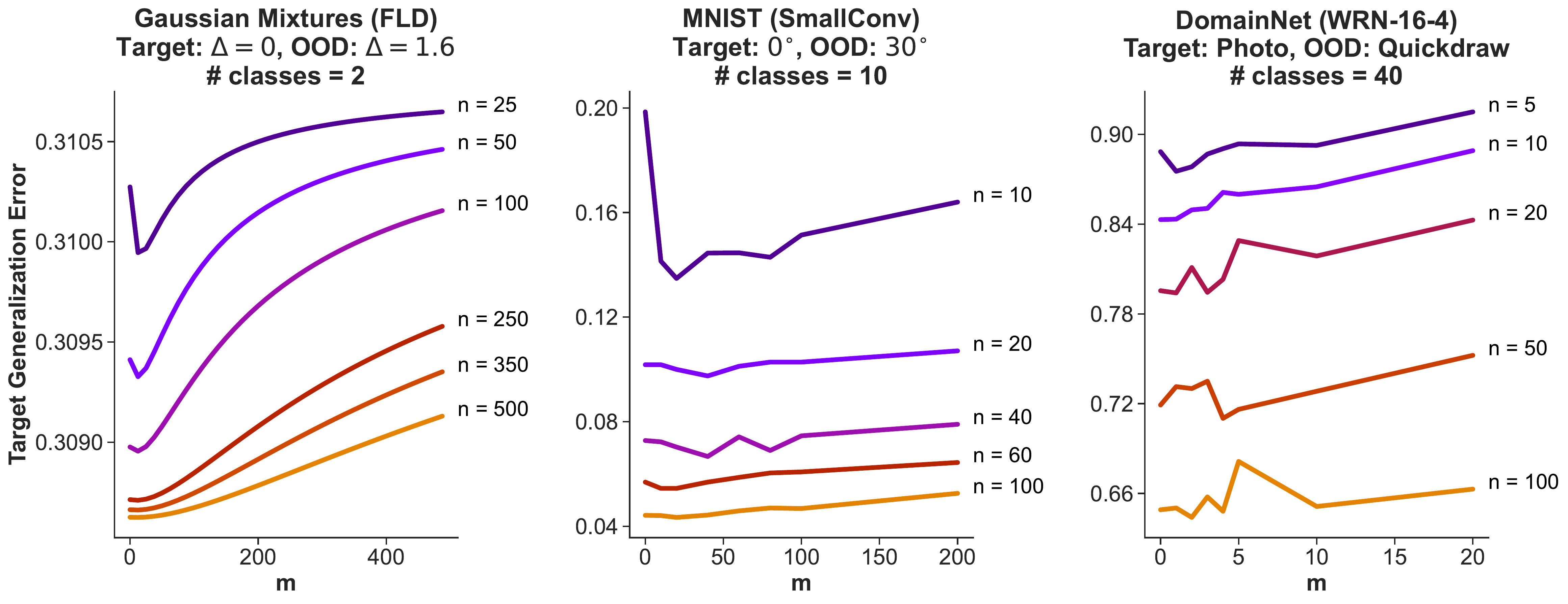}
\caption{We plot the target error (Y-axis) against OOD (X-axis) sample sizes per class $(m)$ for multiple target sample sizes per class $(n)$ across 3 different target-OOD pairs which are: 
\begin{enumerate*}[(1)]
\item a target-OOD pair constructed from a Gaussian mixture model (identical to the one in \cref{s:fld_example}) with $\mu = 5, \sigma= 10$ and OOD translation $\Delta = 1.6$ (left); 
\item the 10-way rotated MNIST classification task where the OOD rotation is $\theta = 30^{\circ}$ (middle) and;
\item a 40-way classification task from DomainNet (see \cref{fig:domainnet-40} for a detailed description) with target and OOD domains of photo and quickdraw respectively (right). 
\end{enumerate*}
We compute the target error analytically for the Gaussian mixture data and compute the empirical average error over 10 and 3 random seeds for the other two distribution pairs respectively. Across all the pairs, we observe non-monotonicity at lower $n$. For larger values of $n$ we believe that the additional OOD samples increase the bias without reducing the variance by much. This could explain why the target error increases monotonically with $m$ at larger values of $n$. 
}
\label{fig:effect-of-target-size}
\vspace{-0.5em}
\end{figure}

\paragraph{Effect of the target sample size on non-monotonicity} Unlike our previous experiments where we fixed the target sample size, in~\cref{fig:effect-of-target-size} we plot the target error as we change both target and OOD sample sizes across 3 different fixed target-OOD pairs. The target generalization error is non-monotonic in the number of OOD samples when we have a small number of target samples for all target-OOD pairs (the solid dark lines that ``dip'' first before increasing later). However, as the number of target samples increases, the non-monotonicity is less pronounced or even completely absent. When we have a large number of target samples, the model is closer to the Bayes error and benefits less from more OOD. Although we do not observe this in~\cref{fig:effect-of-target-size}, we believe that \cref{rem:intuitive} that non-monotonicity could theoretically occur even at large target sample sizes, if the number of samples required to attain the Bayes optimal error is high. 

\section{Can we exploit the non-monotonic trend in the generalization error?}
\label{s:exploit}

\paragraph{Assumption in Sections 3.1 and 3.2}
In the previous section, we discussed non-monotonic trends in generalization error due to the presence of OOD samples in training datasets. If we do not know which samples are OOD, then the generalization for the intended target distribution can deteriorate. 
But it is statistically challenging to identify which samples are OOD; this is discussed in the context of outlier/anomaly detection in~\cref{app:related_work}. We neither propose nor use an explicit method to do this in our paper. Instead, we assume for the sake of analysis that the identities of the target and OOD samples in the datasets are known in advance. We begin by stating the following theorem.

\begin{theorem}[Paraphrased from~\cite{ben2010theory}]
\label{thm:ben_david}
For two distributions $\Pt$ and $\Pout$, let $\hat h_\a$ be the minimizer of the $\a$-weighted empirical loss, i.e.,
\[
    \hat h_\a = \argmin_h \a \hat e_t(h) + (1 - \a) \hat e_o(h)
\]
where $\hat e_t$ and $\hat e_o$ are the empirical losses (see~\cref{eq:task_agnostic_erm}) on $n$ and $m$ training samples drawn from $\Pt$ and $\Pout$, respectively. The generalization error is bounded above by the following inequality
\begin{equation}
\scalemath{0.76}{
e_t(\hat h_{\alpha}) \leq e_t(h_t^*) + 4 \sqrt{ \rbr{\f{\a^2}{n} + \f{(1 - \a)^2}{m}}} \sqrt{V_H - \log \d}\\
    + 2 (1 - \a) d_H(\Pt, \Pout),
} \notag
\end{equation}
with probability at least $1-\d$. Here $h_t^* = \argmin_{h \in H} e_t(h)$ is the target error minimizer; $V_H$ is a constant proportional to the VC-dimension of the hypothesis class $H$ and $d_H(\Pt, \Pout)$ is a notion of relatedness between the distributions $\Pt$ and $\Pout$.
\end{theorem}

In other words, if we use an appropriate value of $\a$ that makes the second and third terms on the right-hand side small, then we can mitigate the deterioration of generalization error due to OOD samples. If the OOD samples are very different from the target samples, i.e., if $d_H(\Pt, \Pout)$ is large, then this theorem suggests that we should pick an $\a \approx 1$. Doing so effectively ignores the OOD samples and the generalization error then decreases monotonically as $\OO(n^{-1/2})$. Note that computation and minimization of the $\alpha$-weighted convex combination of target and OOD losses, $\a \hat e_t(h) + (1 - \a) \hat e_o(h)$, is possible \emph{only} when the identities of target and OOD samples are known in advance.

\subsection{Choosing the optimal $\alpha^*$}

If we define
\(
    \r = \f{\sqrt{V_H - \log \d}}{d_H(\Pt, \Pout)}
\)
to be, roughly speaking, the ratio of the capacity of the hypothesis class and the distance between distributions, then a short calculation shows that for $\a \in [0, 1]$,
\[
\scalemath{0.98}{
    \alpha^* = \begin{cases}
    1 & \text{if } n \geq 4 \r^2, \\
    \f{n}{n + m } \left( 1 + \sqrt{\frac{m^2}{4 \r^2(n + m) - n m }} \right)& \text{else.}\\
\end{cases}
}
\]
This suggests that if we have a hypothesis space with small VC-dimension or if the OOD samples and target samples come from very different distributions, then we should train only on the target samples to obtain optimal error. Otherwise, including the OOD samples after appropriately weighing them using $\a^*$ can give a better generalization error.

It is not easy to estimate $\r$ because it depends upon the VC-dimension of the hypothesis class~\citep{ben2010theory, vedantam2021empirical}. But in general, we can treat $\a$ as a hyperparameter and use validation data to search for its optimal value. For our FLD example we can do slightly better: we can calculate the analytical expression for the generalization error for the hypothesis that minimizes the $\a$-weighted empirical loss (see~\cref{app:weighted_fld,app:weighted_fld_err}) and calculate $\a^*$ by numerically evaluating the expression for $\a \in [0,1]$.

\begin{figure}[h!]
\centering
\includegraphics[width=0.7\linewidth]{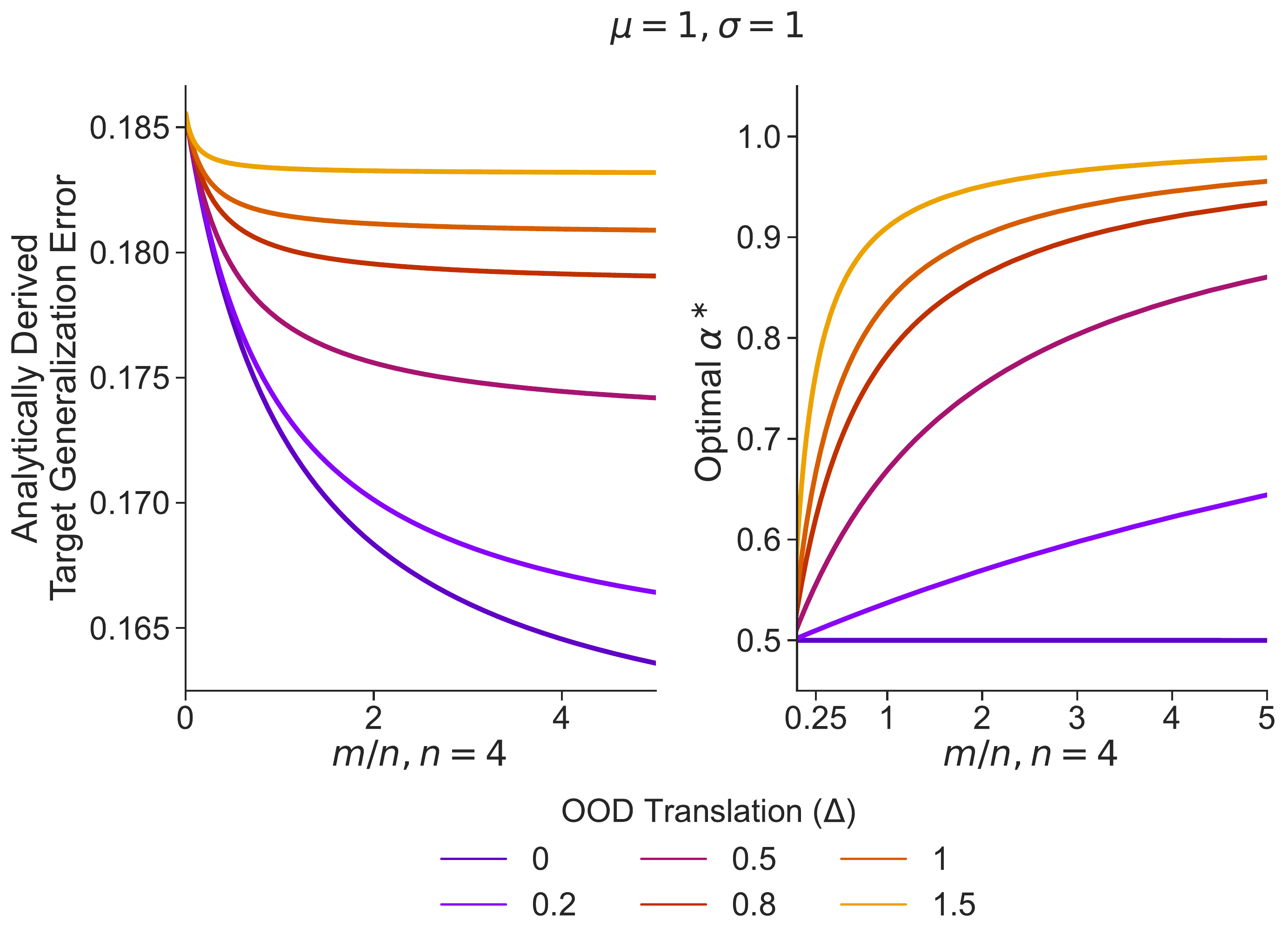}
\caption{\textbf{Left:} Generalization error on the target distribution for the Gaussian mixture model using a weighted objective (\cref{thm:ben_david}) in FLD; see~\cref{app:weighted_fld}. Note that unlike in~\cref{fig:gauss_tasks}, the generalization error monotonically decreases with the number of OOD samples $m$.
\textbf{Right:} The optimal $\a^*$ that yields the smallest target generalization error as a function of the number of OOD samples. Note that $\a^*$ increases as the number of OOD samples $m$ increases; this increase is more drastic for large values of $\D$ and is more gradual for small values of $\D$. Observe that $\a^* = 1/2$ for all values of $m$ if $\D = 0$. See~\cref{app:fld_simualtions} for a numerical simulation.}
\label{fig:weighted_fld_risk}
\end{figure}

\cref{fig:weighted_fld_risk} shows that regardless of the number of OOD samples, $m$, and the relatedness between OOD and target, $\D$, we can obtain a generalization error that is always better than that of a hypothesis trained without OOD samples. In other words, if we choose $\a^*$ appropriately (\cref{fig:gauss_tasks} corresponds to choosing $\a = 1/2$), then we do not suffer from non-monotonic generalization error on the target distribution.

\subsection{Training networks with the $\a$-weighted objective}

\begin{figure}[!t]
\centering
\includegraphics[width=\linewidth]{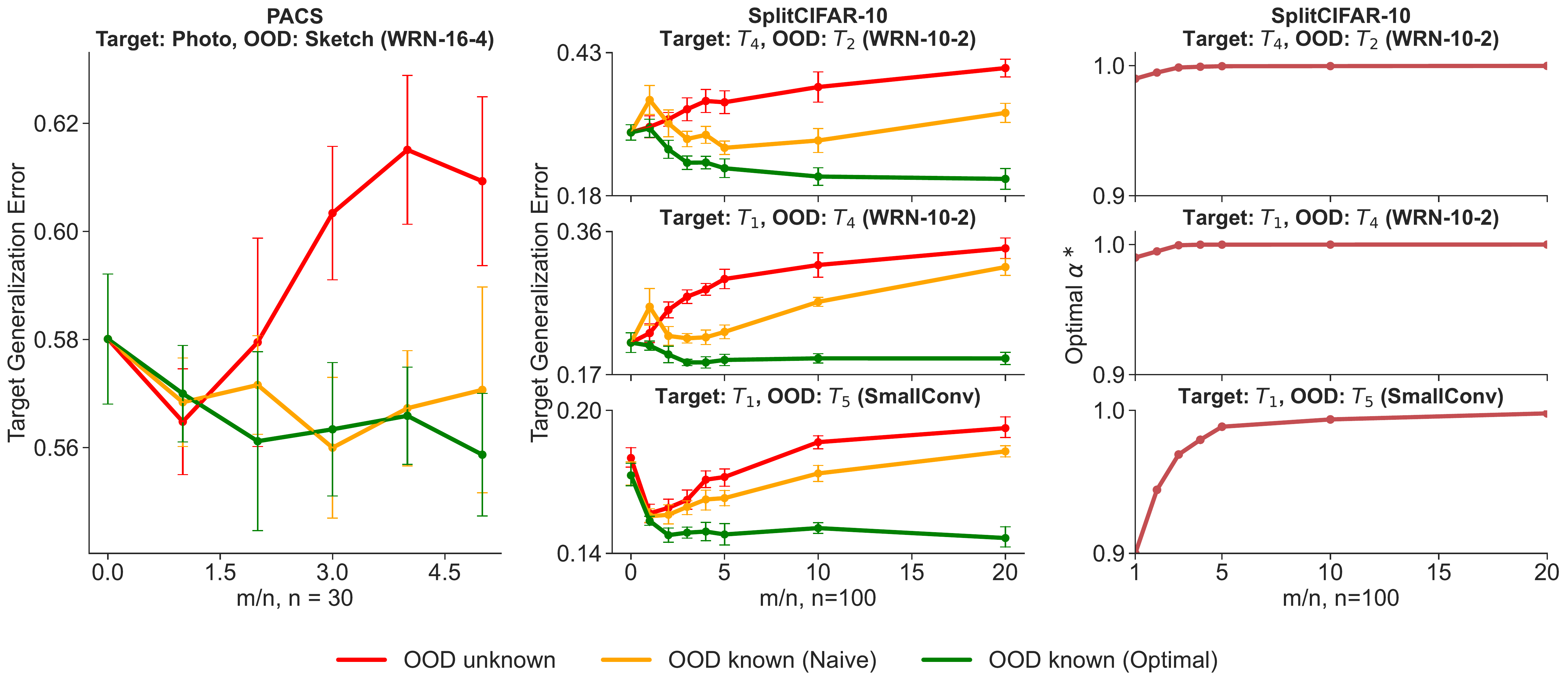}
\caption{
Here we present three settings: minimizing the average loss over target and OOD samples is agnostic to OOD samples present (red), minimizing the sum of the average loss of the target and OOD samples which corresponds to $\a=1/2$ (yellow), minimizing an optimally weighted convex combination of the target and OOD empirical loss (green). The last two settings are only possible when one knows which samples are OOD. For each setting, we plot the generalization error on the target distribution against the number of OOD samples for (target, OOD) pairs from PACS \textbf{(Left)} and CIFAR-10 sub-tasks \textbf{(Middle)}. Unlike in CIFAR-10 task pairs, we observe that in PACS, the target generalization error has a downward trend when $\a = 0.5$ (yellow line, left panel). We speculate that this could be due to the similarity between the target and OOD samples, which causes the model to generalize to the target even at a naive weight.
\textbf{Right:} The optimal $\a^\ast$ obtained via grid search for the three problems in the middle column plotted against different number of OOD samples. The value of $\a^\ast$ lies very close to 1 but it is never exactly 1. In other words, if we use the weighted objective in~\cref{thm:ben_david} then we always obtain some benefit, even if it is marginal when OOD samples are very different from those of the target. Error bars indicate 95\% confidence intervals over 10 experiments.
}
\vspace{-1em}
\label{fig:OOD_known_curves}
\end{figure}

In~\cref{s:non-monotonic_nn}, for a variety of computer vision datasets, we found that for some target-OOD pairs, the generalization error is non-monotonic in the number of OOD samples.
We now show that if we knew which samples were OOD, then we can rectify this trend using an appropriate value of $\a^*$ to weigh the samples differently. In \cref{fig:OOD_known_curves}, we track the test error of the target distribution for three cases: training is agnostic to the presence of OOD samples (red), the learner knows which samples are OOD and uses an $\a=1/2$ in the weighted loss to train (yellow, we call this ``naive''), and when it uses an optimal value of $\a$ using grid-search (green). Searching over $\a$ improves the test error on all these 3 ptarget-OOD pairs.

\begin{figure}[h]
\centering
\includegraphics[width=0.5\linewidth]{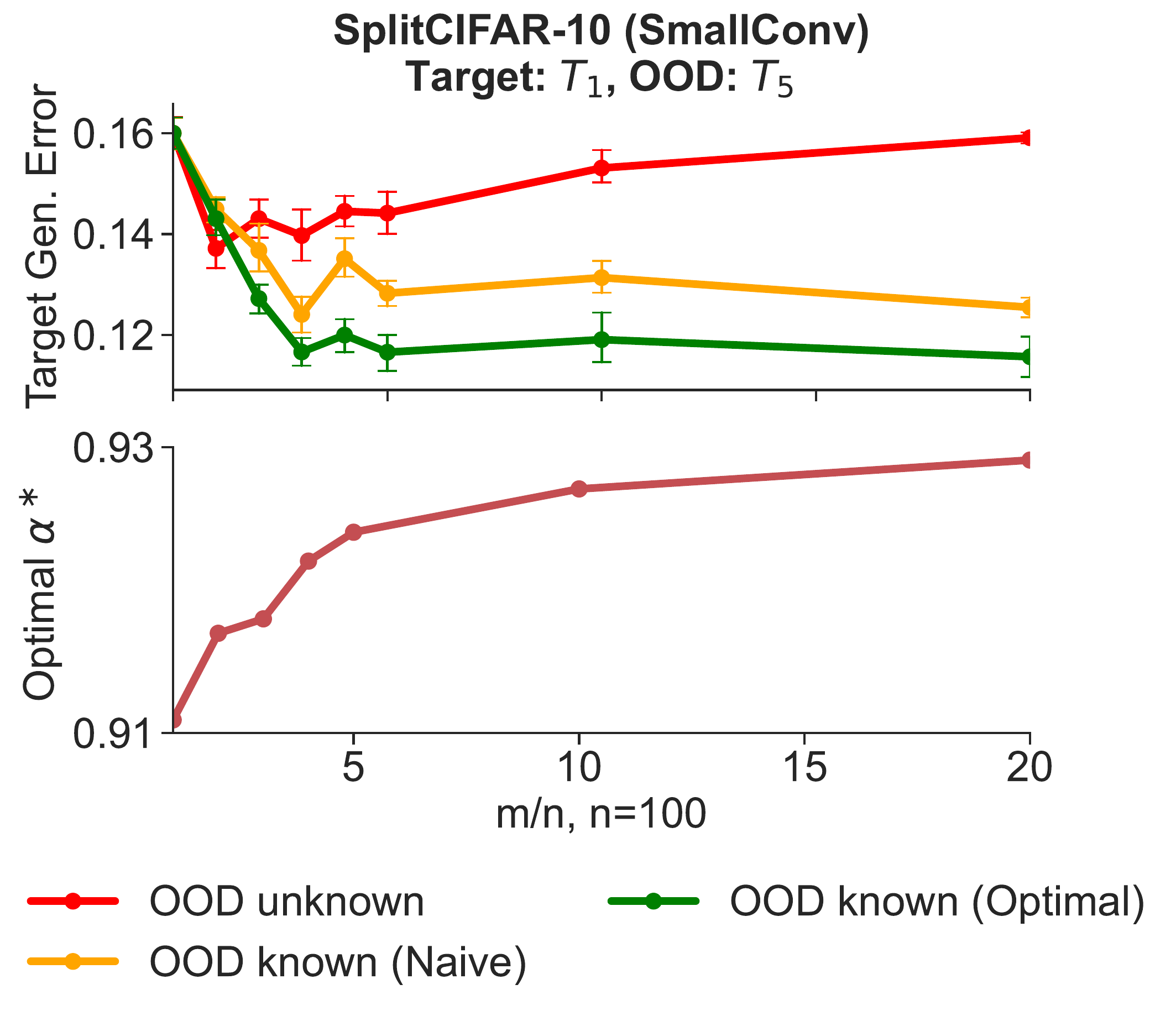}
\vspace*{-1.5em}
\caption{\textbf{Effect of data augmentation} (padding with random cropping and random left/right flipping). Although the network trained in the setting where the OOD sample identities are unknown (red) continues to perform poorly with lots of OOD samples, even a naive weighing of the target and OOD loss ($\a=1/2$) is enough to provide a monotonically decreasing error (yellow) when the OOD sample identities are known. This suggests that data augmentation may mitigate some of the anomalies that arise from OOD data, although we can do better by addressing them specifically using, for instance, the weighted objective (green). Error bars indicate 95\% confidence intervals over 10 experiments.
}
\label{fig:selected_cifar10_ag_op}
\end{figure}

We also conducted another experiment to check if augmentation can help rectify the non-monotonic trend in the generalization error, using the $\a$-weighted objective, i.e., when we know which samples are OOD. As shown in~\cref{fig:selected_cifar10_ag_op}, in this case even naively weighing the objective ($\a=1/2$, yellow) can rectify the non-monotonic trend, using the optimal $\a^*$ (green) further improves the error. This suggests that augmentation is an effective way to mitigate non-monotonic behavior, \emph{but only if we use the $\a$-weighted objective, which requires knowing which samples are OOD.} As we discussed in~\cref{fig:effect-of-techniques}, if we do not know which samples are OOD, then augmentation does not help.

\paragraph{Sampling mini-batches during training}
For $m \gg n$, mini-batches that are sampled uniformly randomly from the dataset will be dominated by OOD samples. As a result, the gradient even if it is still unbiased, is computed using very few target samples. This leads to an increase in the test error, which is particularly noticeable with $\a^*$ chosen appropriately after grid search. We therefore use a biased sampling procedure where each mini-batch contains a fraction $\b$ target samples and the remainder $1-\b$ consists of OOD samples. This parameter controls the bias and variance of the gradient of the target loss ($\b = \frac{n}{n+m}$ gives unbiased gradients with respect to the unweighted total objective and high variance with respect to the target loss when $m \gg n$, see~\cref{sec:app:minibatch}). We found that both $\b = \cbr{0.5, 0.75}$ improve test error.

\paragraph{Weighted objective for over-parameterized networks}
It has been argued previously that weighted objectives are not effective for over-parameterized models such as deep networks because both surrogate losses $\hat e_t(h)$ and $\hat e_o(h)$ are zero when the model fits the training dataset~\citep{byrd2019effect}. It may therefore seem that the weighted objective in~\cref{thm:ben_david} cannot help us mitigate the non-monotonic nature of the generalization error; indeed the minimizer of $\a \hat e_t(h) + (1-\a) \hat e_o(h)$ is the same for any $\a$ if the minimum is exactly zero. Our experiments suggest otherwise: the value of $\a$ does impact the generalization error---even for deep networks. This is perhaps because even if the cross-entropy loss is near-zero for a deep network towards the end of training, it is never exactly zero.

\paragraph{Limitations of the proof-of-concept solution} The numerical and experimental evidence above indicate that even a weighted empirical risk minimization (ERM) algorithm between the target and OOD samples is able to rectify the non-monotonicity. However, this procedure is dependent on two critical ideal conditions: (1) We must know which samples in the dataset are OOD, and (2) We must have a held out dataset of target samples to tune the weight $\alpha$. The difficulty of meeting both of these conditions in reality limits the utility of this procedure as a practical solution to the problem. Instead, we hope that it would serve as a proof-of-concept solution that motivates future research into accurately identifying OOD samples within datasets, designing ways of determining the optimal weights, and developing better procedures for exploiting OOD samples to achieve a lower generalization.

\subsection{Does the upper bound in~\cref{thm:ben_david} inform the non-monotonic trends?}\label{ss:upper_bound_violation}

\cref{thm:ben_david} formed the basis for a proof-of-concept solution in an idealistic setting that exploits OOD samples to reduce target generalization error and effectively correct the non-monotonic trend. Next, we study whether this upper bound predicts the non-monotonic trend.

\begin{figure}[ht!]
    \centering
    \includegraphics[width=0.70\linewidth]{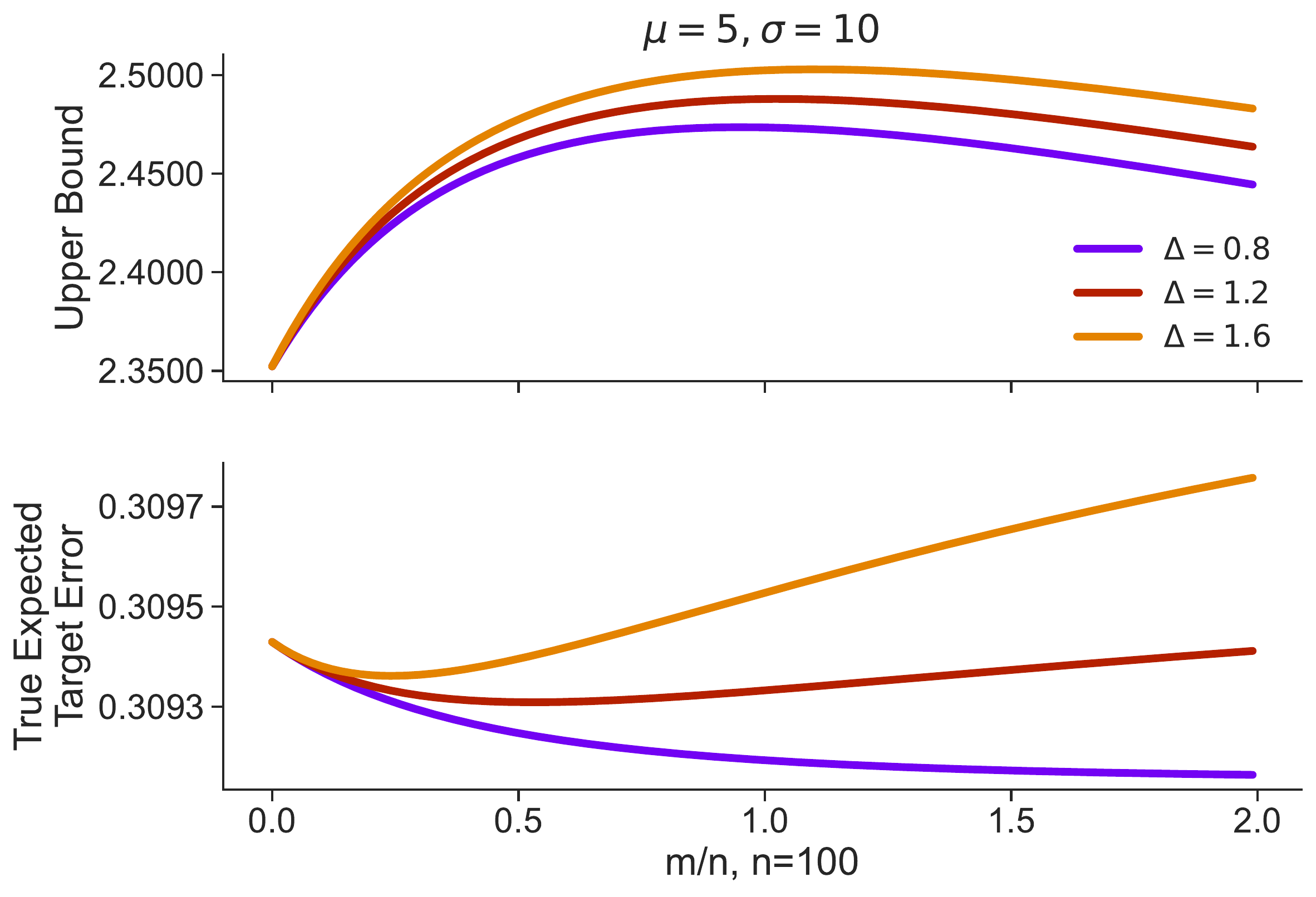}
    \caption{Here we plot the true expected target error (\textbf{bottom}) and the generalization error upper bound value (\textbf{top}) against the $m/n$ ratio for the FLD example ($\mu = 5, \sigma = 10$) in~\cref{fig:gauss_tasks}. The upper bound is significantly vacuous and does not follow the non-monotonic trend of the true target error. However, there are situations when the shape of the upper bound curve is consistent with that of true error (e.g., for large values of shift $\Delta$ between distributions of the target and OOD data). These observations are reported in~\cref{s:app:additional_bound_results}.}
    \label{fig:bound1}
\end{figure}

We return to the setting where we are unaware of the presence of OOD samples in the dataset, and minimize~\cref{eq:task_agnostic_erm}, assuming that all data comes from a single target distribution. We then apply~\cref{thm:ben_david} to our FLD example to derive the following upper bound $U = U(n, m, \Delta)$ for expected error on the target distribution.
\begin{align*}
    \scalemath{0.75}{
    U = \Phi \rbr{-\f \mu \sigma} + 8 \rbr{\frac{\log (\sqrt{32}(n+m+1)/\sqrt{\delta})}{n+m}}^{1/2} + \frac{2m}{n+m}\bigg[ \frac{d^\ast_{H}(\Delta)}{2}  + \lambda \bigg]}
\end{align*}
where $\lambda = \Phi\big(\frac{-\Delta/2 - \mu}{\sigma}\big) + \Phi\big(\frac{\Delta/2 - \mu}{\sigma}\big)$. The derivation (including the procedure of numerically computing $d^\ast_{H}(\Delta)$) is given in the~\cref{s:app:fld_upper_bound}. \cref{fig:bound1} compares the value of the upper bound $U$ with the actual expected target error $e_t(\hat h)$ computed using~\cref{eqn:fld_target_gen_err}.

The upper bound in~\cref{fig:bound1} is vacuous and does not follow a non-monotonic trend when the true error does. Even though its shape fairly agrees with that of true error when $n$ and $\Delta$ are high, it fails to capture the non-monotonic trend we have identified in~\cref{s:fld_example}. The fact that it eludes the grasp of existing theory points to the counter-intuitive nature of this observation and a need for a theoretical investigation of this phenomenon. See~\cref{s:app:additional_bound_results} for more comparisons.

\section{Related Work and Discussion}
\label{app:related_work}

\paragraph{Distribution shift}~\citep{quinonero2008dataset} and its variants such as covariate shift~\citep{ben2012hardness, reddi2015doubly}, concept drift~\citep{mohri2012new, bartlett1992learning, cavallanti2007tracking}, domain shift~\citep{gulrajani2020search, sagawa2021extending, ben2010theory}, sub-population shift~\citep{santurkar2020breeds, hu2018does, sagawa2019distributionally}, data poisoning~\citep{yang2017generative, steinhardt2017certified}, geometric and semantic nuisances~\citep{van2019towards}, and flawed annotations~\citep{frenay2013classification} can lead to the presence of OOD samples in a curated dataset, and thereby may yield sub-optimal generalization error on the desired task. While these problems have been studied in the sense of an out-of-domain \emph{distribution}, we believe that we have identified a fundamentally different phenomenon, namely a non-monotonic trend in the generalization error with respect to the OOD samples in training data.

\vspace*{-0.5em}
\paragraph{Internal Dataset Shift} A recent body of works~\citep{kaplun2022deconstructing, swayamdipta2020dataset, siddiqui2022metadata, jain2022data, maini2022characterizing} has investigated the presence of noisy, hard-to-learn, and/or negatively influential samples in popular vision benchmarks. Existence of such OOD samples indicates that the internal dataset shift may be a widespread problem in real datasets. Such circumstances may give rise to undesired non-monotonic trends in generalization error, as we have described in our work. 

\vspace*{-0.5em}
\paragraph{Domain Adaptation}  While most works listed above provide attractive ways of adapting or being robust to various modes of shift, a part of our work addresses the question: \emph{if} we know which samples are OOD, then can we optimally utilize them to achieve a better generalization on the desired target task? This is related to domain adaptation~\citep{ben2010theory, mansour2008domain, pan2010domain, ganin2016domain, cortes2019adaptation}. A large body of work uses weighted-ERM based methods for domain adaptation~\citep{ben2010theory, zhang2012generalization, blitzer2007learning, bu2022characterizing, hanneke2019value, redko2017theoretical, wang2019transfer, ben2006analysis}; this is either done to address domain shift or to address different distributions of tasks in a transfer or multi-task learning setting. This body of work is of interest for us, except that in our case, the ``source'' task is actually the OOD samples.

\vspace*{-0.5em}
\paragraph{Connection with the theory of domain adaptation}
While generalization bounds for weighted-ERM like those of~\citet{ben2010theory} are understood to be meaningful (if not tight; see~\citet{vedantam2021empirical}) for large sample sizes, our work identifies an unusual non-monotonic trend in the generalization error of the target task. Note that the upper bound proposed by~\citet{ben2010theory} can be used when we do not know the identity of the OOD samples by setting $\a = \f{n}{n+m}$. However, our experiments in~\cref{ss:upper_bound_violation} reveal that this bound is significantly vacuous and does not predict the non-monotonic trends we have identified. There is another discrepancy here, e.g., we notice that the upper bound for naively weighted empirical error ($\a = 1/2$) does not have a non-monotonic trend. A more recent paper by~\citet{bu2022characterizing} presents an exact characterization of the target generalization error using conditional symmetrized Kullback-Leibler information between the output hypothesis and target samples given the source samples. While they do not identify non-monotonic trends in target generalization error, their tools can potentially be useful to characterize the phenomenon discovered in our work.

\vspace*{-0.5em}
\paragraph{Domain Generalization} seeks to learn a predictor from multiple domains that could perform well on some \emph{unseen} test domain. This unseen test domain can be thought as OOD data. Since no training data is available during the training, the learner needs to make some additional assumptions; one popular assumption is to learn invariances across training and testing domains~\citep{gulrajani2020search, arjovsky2019invariant, sun2016deep}. We use several benchmark datasets from this literature, but the goals of this body of work and ours are very different because we are interested only in generalizing on the target task, not generalizing to the domain of the OOD samples.

\vspace*{-0.5em}
\paragraph{Outlier and OOD Detection} Identifying OOD samples within a dataset \emph{prior to training} can be thought of as a variation of the outlier detection (OD) problem~\citep{ben2010outlier, boukerche2020outlier, wang2019progress, fischler1981random}. These methods aim to detect outliers by searching for the model fitted by the majority of samples. But this remains a largely unsolved problem for high-dimensional data~\citep{thudumu2020comprehensive}. Another related but different problem is ``OOD detection''~\citep{ren2019likelihood, winkens2020contrastive, fort2021exploring, liu2020energy} which focuses on detecting data that is different from what was used for training (also see the works of \citet{ming2022impact, sun2022out} who demonstrate that certain detected OOD samples can turn out to be semantically similar to training samples). 



\vspace*{-0.5em}
\section{Acknowledgements}

ADS and JTV were supported by the NSF AI Institute Planning award (\#2020312), NSF-Simons Research Collaborations on the Mathematical and Scientific Foundations of Deep Learning (MoDL) and THEORINET (\#2031985). RR and PC were supported by grants from the National Science Foundation (IIS-2145164, CCF-2212519), Office of Naval Research (N00014-22-1-2255), and cloud computing credits from Amazon Web Services.

\clearpage
\bibliographystyle{icml2023}
\bibliography{references}

\begin{thebibliography}{56}
\providecommand{\natexlab}[1]{#1}
\providecommand{\url}[1]{\texttt{#1}}
\expandafter\ifx\csname urlstyle\endcsname\relax
  \providecommand{\doi}[1]{doi: #1}\else
  \providecommand{\doi}{doi: \begingroup \urlstyle{rm}\Url}\fi

\bibitem[Arjovsky et~al.(2019)Arjovsky, Bottou, Gulrajani, and
  Lopez-Paz]{arjovsky2019invariant}
Arjovsky, M., Bottou, L., Gulrajani, I., and Lopez-Paz, D.
\newblock Invariant risk minimization.
\newblock \emph{arXiv preprint arXiv:1907.02893}, 2019.

\bibitem[Bartlett(1992)]{bartlett1992learning}
Bartlett, P.~L.
\newblock Learning with a slowly changing distribution.
\newblock In \emph{Proceedings of the fifth annual workshop on Computational
  learning theory}, pp.\  243--252, 1992.

\bibitem[Ben-David \& Urner(2012)Ben-David and Urner]{ben2012hardness}
Ben-David, S. and Urner, R.
\newblock On the hardness of domain adaptation and the utility of unlabeled
  target samples.
\newblock In \emph{International Conference on Algorithmic Learning Theory},
  pp.\  139--153. Springer, 2012.

\bibitem[Ben-David et~al.(2006)Ben-David, Blitzer, Crammer, and
  Pereira]{ben2006analysis}
Ben-David, S., Blitzer, J., Crammer, K., and Pereira, F.
\newblock Analysis of representations for domain adaptation.
\newblock \emph{Advances in neural information processing systems}, 19, 2006.

\bibitem[Ben-David et~al.(2010)Ben-David, Blitzer, Crammer, Kulesza, Pereira,
  and Vaughan]{ben2010theory}
Ben-David, S., Blitzer, J., Crammer, K., Kulesza, A., Pereira, F., and Vaughan,
  J.~W.
\newblock A theory of learning from different domains.
\newblock \emph{Machine learning}, 79\penalty0 (1):\penalty0 151--175, 2010.

\bibitem[Ben-Gal(2010)]{ben2010outlier}
Ben-Gal, I.
\newblock Outlier detection.
\newblock \emph{Data mining and knowledge discovery handbook}, pp.\  117--130,
  2010.

\bibitem[Blitzer et~al.(2007)Blitzer, Crammer, Kulesza, Pereira, and
  Wortman]{blitzer2007learning}
Blitzer, J., Crammer, K., Kulesza, A., Pereira, F., and Wortman, J.
\newblock Learning bounds for domain adaptation.
\newblock \emph{Advances in neural information processing systems}, 20, 2007.

\bibitem[Boukerche et~al.(2020)Boukerche, Zheng, and
  Alfandi]{boukerche2020outlier}
Boukerche, A., Zheng, L., and Alfandi, O.
\newblock Outlier detection: Methods, models, and classification.
\newblock \emph{ACM Computing Surveys (CSUR)}, 53\penalty0 (3):\penalty0 1--37,
  2020.

\bibitem[Bu et~al.(2022)Bu, Aminian, Toni, Wornell, and
  Rodrigues]{bu2022characterizing}
Bu, Y., Aminian, G., Toni, L., Wornell, G.~W., and Rodrigues, M.
\newblock Characterizing and understanding the generalization error of transfer
  learning with gibbs algorithm.
\newblock In \emph{International Conference on Artificial Intelligence and
  Statistics}, pp.\  8673--8699. PMLR, 2022.

\bibitem[Byrd \& Lipton(2019)Byrd and Lipton]{byrd2019effect}
Byrd, J. and Lipton, Z.
\newblock What is the effect of importance weighting in deep learning?
\newblock In \emph{{International Conference on Machine Learning}}, pp.\
  872--881, 2019.

\bibitem[Cavallanti et~al.(2007)Cavallanti, Cesa-Bianchi, and
  Gentile]{cavallanti2007tracking}
Cavallanti, G., Cesa-Bianchi, N., and Gentile, C.
\newblock Tracking the best hyperplane with a simple budget perceptron.
\newblock \emph{Machine Learning}, 69\penalty0 (2):\penalty0 143--167, 2007.

\bibitem[Cortes et~al.(2019)Cortes, Mohri, and Medina]{cortes2019adaptation}
Cortes, C., Mohri, M., and Medina, A.~M.
\newblock Adaptation based on generalized discrepancy.
\newblock \emph{The Journal of Machine Learning Research}, 20\penalty0
  (1):\penalty0 1--30, 2019.

\bibitem[Darlow et~al.(2018)Darlow, Crowley, Antoniou, and
  Storkey]{darlow2018cinic}
Darlow, L.~N., Crowley, E.~J., Antoniou, A., and Storkey, A.~J.
\newblock Cinic-10 is not imagenet or cifar-10.
\newblock \emph{arXiv preprint arXiv:1810.03505}, 2018.

\bibitem[Fischler \& Bolles(1981)Fischler and Bolles]{fischler1981random}
Fischler, M.~A. and Bolles, R.~C.
\newblock Random sample consensus: a paradigm for model fitting with
  applications to image analysis and automated cartography.
\newblock \emph{Communications of the ACM}, 24\penalty0 (6):\penalty0 381--395,
  1981.

\bibitem[Fort et~al.(2021)Fort, Ren, and Lakshminarayanan]{fort2021exploring}
Fort, S., Ren, J., and Lakshminarayanan, B.
\newblock Exploring the limits of out-of-distribution detection.
\newblock \emph{Advances in Neural Information Processing Systems},
  34:\penalty0 7068--7081, 2021.

\bibitem[Fr{\'e}nay \& Verleysen(2013)Fr{\'e}nay and
  Verleysen]{frenay2013classification}
Fr{\'e}nay, B. and Verleysen, M.
\newblock Classification in the presence of label noise: a survey.
\newblock \emph{IEEE transactions on neural networks and learning systems},
  25\penalty0 (5):\penalty0 845--869, 2013.

\bibitem[Ganin et~al.(2016)Ganin, Ustinova, Ajakan, Germain, Larochelle,
  Laviolette, Marchand, and Lempitsky]{ganin2016domain}
Ganin, Y., Ustinova, E., Ajakan, H., Germain, P., Larochelle, H., Laviolette,
  F., Marchand, M., and Lempitsky, V.
\newblock Domain-adversarial training of neural networks.
\newblock \emph{The journal of machine learning research}, 17\penalty0
  (1):\penalty0 2096--2030, 2016.

\bibitem[Ghifary et~al.(2015)Ghifary, Kleijn, Zhang, and
  Balduzzi]{ghifary2015domain}
Ghifary, M., Kleijn, W.~B., Zhang, M., and Balduzzi, D.
\newblock Domain generalization for object recognition with multi-task
  autoencoders.
\newblock In \emph{Proceedings of the IEEE international conference on computer
  vision}, pp.\  2551--2559, 2015.

\bibitem[Gulrajani \& Lopez-Paz(2020)Gulrajani and
  Lopez-Paz]{gulrajani2020search}
Gulrajani, I. and Lopez-Paz, D.
\newblock In search of lost domain generalization.
\newblock \emph{arXiv preprint arXiv:2007.01434}, 2020.

\bibitem[Hanneke \& Kpotufe(2019)Hanneke and Kpotufe]{hanneke2019value}
Hanneke, S. and Kpotufe, S.
\newblock On the value of target data in transfer learning.
\newblock \emph{Advances in Neural Information Processing Systems}, 32, 2019.

\bibitem[Hu et~al.(2018)Hu, Niu, Sato, and Sugiyama]{hu2018does}
Hu, W., Niu, G., Sato, I., and Sugiyama, M.
\newblock Does distributionally robust supervised learning give robust
  classifiers?
\newblock In \emph{International Conference on Machine Learning}, pp.\
  2029--2037. PMLR, 2018.

\bibitem[Jain et~al.(2022)Jain, Salman, Khaddaj, Wong, Park, and
  Madry]{jain2022data}
Jain, S., Salman, H., Khaddaj, A., Wong, E., Park, S.~M., and Madry, A.
\newblock A data-based perspective on transfer learning.
\newblock \emph{arXiv preprint arXiv:2207.05739}, 2022.

\bibitem[Kaplun et~al.(2022)Kaplun, Ghosh, Garg, Barak, and
  Nakkiran]{kaplun2022deconstructing}
Kaplun, G., Ghosh, N., Garg, S., Barak, B., and Nakkiran, P.
\newblock Deconstructing distributions: A pointwise framework of learning.
\newblock \emph{arXiv preprint arXiv:2202.09931}, 2022.

\bibitem[Kumar et~al.(2022)Kumar, Raghunathan, Jones, Ma, and
  Liang]{kumar2022fine}
Kumar, A., Raghunathan, A., Jones, R., Ma, T., and Liang, P.
\newblock Fine-tuning can distort pretrained features and underperform
  out-of-distribution.
\newblock \emph{arXiv preprint arXiv:2202.10054}, 2022.

\bibitem[Li et~al.(2017)Li, Yang, Song, and Hospedales]{li2017deeper}
Li, D., Yang, Y., Song, Y.-Z., and Hospedales, T.~M.
\newblock Deeper, broader and artier domain generalization.
\newblock In \emph{Proceedings of the IEEE international conference on computer
  vision}, pp.\  5542--5550, 2017.

\bibitem[Liaw et~al.(2018)Liaw, Liang, Nishihara, Moritz, Gonzalez, and
  Stoica]{liaw2018tune}
Liaw, R., Liang, E., Nishihara, R., Moritz, P., Gonzalez, J.~E., and Stoica, I.
\newblock Tune: A research platform for distributed model selection and
  training.
\newblock \emph{arXiv preprint arXiv:1807.05118}, 2018.

\bibitem[Liu et~al.(2020)Liu, Wang, Owens, and Li]{liu2020energy}
Liu, W., Wang, X., Owens, J., and Li, Y.
\newblock Energy-based out-of-distribution detection.
\newblock \emph{Advances in neural information processing systems},
  33:\penalty0 21464--21475, 2020.

\bibitem[Maini et~al.(2022)Maini, Garg, Lipton, and
  Kolter]{maini2022characterizing}
Maini, P., Garg, S., Lipton, Z.~C., and Kolter, J.~Z.
\newblock Characterizing datapoints via second-split forgetting.
\newblock \emph{arXiv preprint arXiv:2210.15031}, 2022.

\bibitem[Mansour et~al.(2008)Mansour, Mohri, and
  Rostamizadeh]{mansour2008domain}
Mansour, Y., Mohri, M., and Rostamizadeh, A.
\newblock Domain adaptation with multiple sources.
\newblock \emph{Advances in neural information processing systems}, 21, 2008.

\bibitem[Ming et~al.(2022)Ming, Yin, and Li]{ming2022impact}
Ming, Y., Yin, H., and Li, Y.
\newblock On the impact of spurious correlation for out-of-distribution
  detection.
\newblock In \emph{Proceedings of the AAAI Conference on Artificial
  Intelligence}, volume~36, pp.\  10051--10059, 2022.

\bibitem[Mohri \& Mu{\~n}oz~Medina(2012)Mohri and
  Mu{\~n}oz~Medina]{mohri2012new}
Mohri, M. and Mu{\~n}oz~Medina, A.
\newblock New analysis and algorithm for learning with drifting distributions.
\newblock In \emph{International Conference on Algorithmic Learning Theory},
  pp.\  124--138. Springer, 2012.

\bibitem[Pan et~al.(2010)Pan, Tsang, Kwok, and Yang]{pan2010domain}
Pan, S.~J., Tsang, I.~W., Kwok, J.~T., and Yang, Q.
\newblock Domain adaptation via transfer component analysis.
\newblock \emph{IEEE transactions on neural networks}, 22\penalty0
  (2):\penalty0 199--210, 2010.

\bibitem[Peng et~al.(2019)Peng, Bai, Xia, Huang, Saenko, and
  Wang]{peng2019moment}
Peng, X., Bai, Q., Xia, X., Huang, Z., Saenko, K., and Wang, B.
\newblock Moment matching for multi-source domain adaptation.
\newblock In \emph{Proceedings of the IEEE/CVF international conference on
  computer vision}, pp.\  1406--1415, 2019.

\bibitem[Quinonero-Candela et~al.(2008)Quinonero-Candela, Sugiyama,
  Schwaighofer, and Lawrence]{quinonero2008dataset}
Quinonero-Candela, J., Sugiyama, M., Schwaighofer, A., and Lawrence, N.~D.
\newblock \emph{Dataset shift in machine learning}.
\newblock Mit Press, 2008.

\bibitem[Reddi et~al.(2015)Reddi, Poczos, and Smola]{reddi2015doubly}
Reddi, S., Poczos, B., and Smola, A.
\newblock Doubly robust covariate shift correction.
\newblock In \emph{Proceedings of the AAAI Conference on Artificial
  Intelligence}, volume~29, 2015.

\bibitem[Redko et~al.(2017)Redko, Habrard, and Sebban]{redko2017theoretical}
Redko, I., Habrard, A., and Sebban, M.
\newblock Theoretical analysis of domain adaptation with optimal transport.
\newblock In \emph{Joint European Conference on Machine Learning and Knowledge
  Discovery in Databases}, pp.\  737--753. Springer, 2017.

\bibitem[Ren et~al.(2019)Ren, Liu, Fertig, Snoek, Poplin, Depristo, Dillon, and
  Lakshminarayanan]{ren2019likelihood}
Ren, J., Liu, P.~J., Fertig, E., Snoek, J., Poplin, R., Depristo, M., Dillon,
  J., and Lakshminarayanan, B.
\newblock Likelihood ratios for out-of-distribution detection.
\newblock \emph{Advances in neural information processing systems}, 32, 2019.

\bibitem[Sagawa et~al.(2019)Sagawa, Koh, Hashimoto, and
  Liang]{sagawa2019distributionally}
Sagawa, S., Koh, P.~W., Hashimoto, T.~B., and Liang, P.
\newblock Distributionally robust neural networks for group shifts: On the
  importance of regularization for worst-case generalization.
\newblock \emph{arXiv preprint arXiv:1911.08731}, 2019.

\bibitem[Sagawa et~al.(2021)Sagawa, Koh, Lee, Gao, Xie, Shen, Kumar, Hu,
  Yasunaga, Marklund, et~al.]{sagawa2021extending}
Sagawa, S., Koh, P.~W., Lee, T., Gao, I., Xie, S.~M., Shen, K., Kumar, A., Hu,
  W., Yasunaga, M., Marklund, H., et~al.
\newblock Extending the wilds benchmark for unsupervised adaptation.
\newblock \emph{arXiv preprint arXiv:2112.05090}, 2021.

\bibitem[Santurkar et~al.(2020)Santurkar, Tsipras, and
  Madry]{santurkar2020breeds}
Santurkar, S., Tsipras, D., and Madry, A.
\newblock Breeds: Benchmarks for subpopulation shift.
\newblock \emph{arXiv preprint arXiv:2008.04859}, 2020.

\bibitem[Siddiqui et~al.(2022)Siddiqui, Rajkumar, Maharaj, Krueger, and
  Hooker]{siddiqui2022metadata}
Siddiqui, S.~A., Rajkumar, N., Maharaj, T., Krueger, D., and Hooker, S.
\newblock Metadata archaeology: Unearthing data subsets by leveraging training
  dynamics.
\newblock \emph{arXiv preprint arXiv:2209.10015}, 2022.

\bibitem[Smola \& Sch{\"o}lkopf(1998)Smola and
  Sch{\"o}lkopf]{smola1998learning}
Smola, A.~J. and Sch{\"o}lkopf, B.
\newblock \emph{Learning with kernels}, volume~4.
\newblock 1998.

\bibitem[Steinhardt et~al.(2017)Steinhardt, Koh, and
  Liang]{steinhardt2017certified}
Steinhardt, J., Koh, P. W.~W., and Liang, P.~S.
\newblock Certified defenses for data poisoning attacks.
\newblock \emph{Advances in neural information processing systems}, 30, 2017.

\bibitem[Sun \& Saenko(2016)Sun and Saenko]{sun2016deep}
Sun, B. and Saenko, K.
\newblock Deep coral: Correlation alignment for deep domain adaptation.
\newblock In \emph{European conference on computer vision}, pp.\  443--450.
  Springer, 2016.

\bibitem[Sun et~al.(2022)Sun, Ming, Zhu, and Li]{sun2022out}
Sun, Y., Ming, Y., Zhu, X., and Li, Y.
\newblock Out-of-distribution detection with deep nearest neighbors.
\newblock In \emph{International Conference on Machine Learning}, pp.\
  20827--20840. PMLR, 2022.

\bibitem[Swayamdipta et~al.(2020)Swayamdipta, Schwartz, Lourie, Wang,
  Hajishirzi, Smith, and Choi]{swayamdipta2020dataset}
Swayamdipta, S., Schwartz, R., Lourie, N., Wang, Y., Hajishirzi, H., Smith,
  N.~A., and Choi, Y.
\newblock Dataset cartography: Mapping and diagnosing datasets with training
  dynamics.
\newblock \emph{arXiv preprint arXiv:2009.10795}, 2020.

\bibitem[Thudumu et~al.(2020)Thudumu, Branch, Jin, and
  Singh]{thudumu2020comprehensive}
Thudumu, S., Branch, P., Jin, J., and Singh, J.
\newblock A comprehensive survey of anomaly detection techniques for high
  dimensional big data.
\newblock \emph{Journal of Big Data}, 7:\penalty0 1--30, 2020.

\bibitem[Van~Horn(2019)]{van2019towards}
Van~Horn, G.~R.
\newblock \emph{Towards a Visipedia: Combining Computer Vision and Communities
  of Experts}.
\newblock PhD thesis, California Institute of Technology, 2019.

\bibitem[Vedantam et~al.(2021)Vedantam, Lopez-Paz, and
  Schwab]{vedantam2021empirical}
Vedantam, R., Lopez-Paz, D., and Schwab, D.~J.
\newblock An empirical investigation of domain generalization with empirical
  risk minimizers.
\newblock \emph{Advances in Neural Information Processing Systems},
  34:\penalty0 28131--28143, 2021.

\bibitem[Wang et~al.(2019{\natexlab{a}})Wang, Mendez, Cai, and
  Eaton]{wang2019transfer}
Wang, B., Mendez, J., Cai, M., and Eaton, E.
\newblock Transfer learning via minimizing the performance gap between domains.
\newblock \emph{Advances in Neural Information Processing Systems}, 32,
  2019{\natexlab{a}}.

\bibitem[Wang et~al.(2019{\natexlab{b}})Wang, Bah, and
  Hammad]{wang2019progress}
Wang, H., Bah, M.~J., and Hammad, M.
\newblock Progress in outlier detection techniques: A survey.
\newblock \emph{Ieee Access}, 7:\penalty0 107964--108000, 2019{\natexlab{b}}.

\bibitem[Winkens et~al.(2020)Winkens, Bunel, Roy, Stanforth, Natarajan, Ledsam,
  MacWilliams, Kohli, Karthikesalingam, Kohl, et~al.]{winkens2020contrastive}
Winkens, J., Bunel, R., Roy, A.~G., Stanforth, R., Natarajan, V., Ledsam,
  J.~R., MacWilliams, P., Kohli, P., Karthikesalingam, A., Kohl, S., et~al.
\newblock Contrastive training for improved out-of-distribution detection.
\newblock \emph{arXiv preprint arXiv:2007.05566}, 2020.

\bibitem[Yang et~al.(2017)Yang, Wu, Li, and Chen]{yang2017generative}
Yang, C., Wu, Q., Li, H., and Chen, Y.
\newblock Generative poisoning attack method against neural networks.
\newblock \emph{arXiv preprint arXiv:1703.01340}, 2017.

\bibitem[Zagoruyko \& Komodakis(2016)Zagoruyko and
  Komodakis]{zagoruyko2016wide}
Zagoruyko, S. and Komodakis, N.
\newblock Wide residual networks.
\newblock \emph{arXiv preprint arXiv:1605.07146}, 2016.

\bibitem[Zenke et~al.(2017)Zenke, Poole, and Ganguli]{zenke_continual_2017}
Zenke, F., Poole, B., and Ganguli, S.
\newblock Continual {Learning} {Through} {Synaptic} {Intelligence}.
\newblock In \emph{{International Conference on Machine Learning}}, pp.\
  3987--3995, 2017.

\bibitem[Zhang et~al.(2012)Zhang, Zhang, and Ye]{zhang2012generalization}
Zhang, C., Zhang, L., and Ye, J.
\newblock Generalization bounds for domain adaptation.
\newblock \emph{Advances in neural information processing systems}, 25, 2012.

\end{thebibliography}


\renewcommand\thefigure{A\arabic{figure}}
\renewcommand\thetable{A\arabic{table}}
\setcounter{figure}{0}
\setcounter{table}{0}

\clearpage
\onecolumn
\appendix

\section{Fisher's Linear Discriminant (FLD)}

\subsection{Synthetic Datasets}
\label{sec:app:synthetic}

The target data is sampled from the distribution $P_t$ and the OOD data is sampled from the distribution $P_o$; Both distributions have two classes and one-dimensional inputs. In both distrbutions, each class is sampled from a univariate Gaussian distribution. The distribution of the OOD data is the target distribution translated by $\Delta$. In summary, the target distribution has the class conditional densities,
\begin{align*}
    f_{t, 0}
        &\stackrel{d}{=} \mathcal{N}(-\mu, \sigma^2) \\
    f_{t, 1}
        &\stackrel{d}{=} \mathcal{N}(+\mu, \sigma^2),
\end{align*}
while the OOD distribution has the class conditional densities,
\begin{align*}
    f_{o, 0}
        &\stackrel{d}{=} \mathcal{N}(\Delta -\mu, \sigma^2) \\
    f_{o, 1}
        &\stackrel{d}{=} \mathcal{N}(\Delta + \mu, \sigma^2).
\end{align*}
We also assume that both the target and OOD distributions have the same label distribution with equal class prior probabilities, i.e. $p(y_t=1) = p(y_o=1) = \pi = \frac{1}{2}$. \cref{fig:gauss_tasks} (left) depicts $\Pt$ and $\Pout$ pictorially.
\vspace{0.5em}
\begin{figure}[!h]
\centering
\includegraphics[width=0.4\linewidth]{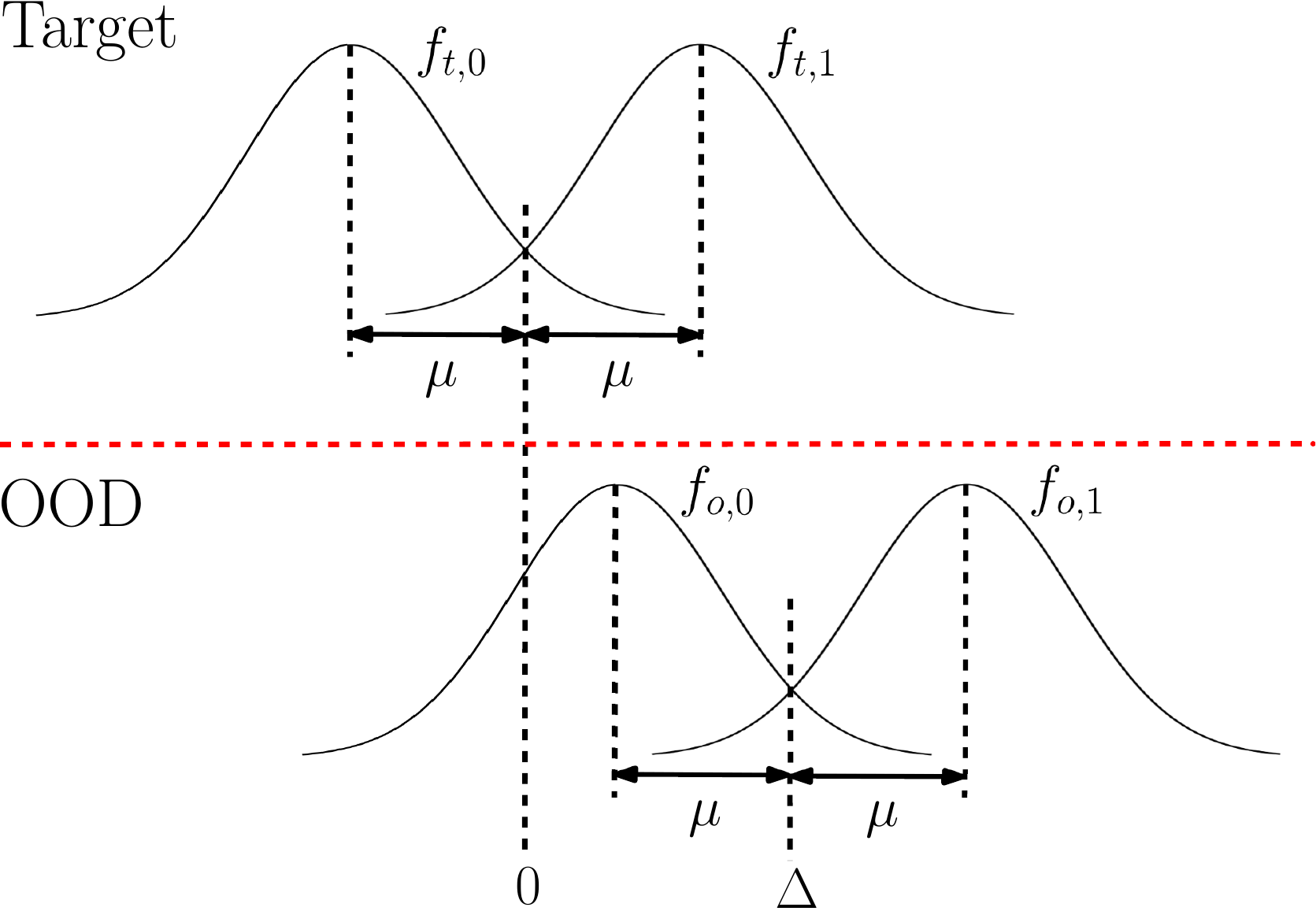}
\caption{A picture of synthetic target and OOD distributions.}
\end{figure}

\subsection{OOD-Agnostic Fisher's Linear Discriminant}\label{app:fld_derivation}

In this section, we derive FLD when we have samples from a single distribution -- which is also applicable to the OOD-agnostic (when the identity of the OOD samples are not known) setting. Consider a binary classification problem with $D_t = \{(x_i, y_i)\}_{i=1}^{n} \sim P_t$ where $x_i \in X \subseteq \mathbb{R}^d$ and $y_i \in Y = \{0, 1\}$.

Let $f_k$ and $\pi_k$ be the conditional density and prior probability of class $k$ ($k \in \{0, 1\})$ respectively. The probability that $x$ belongs to class $k$ is
\begin{equation}
    p(y = k\mid x) = \frac{\pi_k f_k(x)}{\pi_0 f_0(x) + \pi_1 f_1(x)} \notag,
\end{equation}
and the maximum \emph{a posteriori} estimate of the class label is
\begin{equation}\label{eqn:classifier}
    h(x) = \argmax_{k \in \{0, 1\}} \ p(y = k \mid x) = \argmax_{k \in \{0, 1\}} \ \log (\pi_k f_k(x)).
\end{equation}

Fisher's linear discriminant (FLD) assumes that each $f_k$ is a multivariate Gaussian distribution with the same covariance matrix
$\Sigma$, i.e,
\begin{equation}
    f_k(x) = \frac{1}{(2\pi)^{d/2}|\Sigma|^{1/2}} \exp \bigg( -\frac{1}{2} (x-\mu_k)^\top \Sigma^{-1}(x-\mu_k) \bigg). \notag 
\end{equation}
Under this assumption, the joint-density $f$ of $(x, y)$ becomes,
\begin{equation}
    f(x, y) \propto \prod_{k=0}^{1} \bigg[ \frac{\pi_k}{|\Sigma|^{1/2}}\exp \bigg( -\frac{1}{2} (x-\mu_k)^\top \Sigma^{-1}(x-\mu_k) \bigg) \bigg]^{\mathbbm{1}[y=k]} \notag
\end{equation}
Therefore, the log-likelihood $l(\mu_0, \mu_1, \Sigma, \pi_0, \pi_1)$ over $D_t$ is given by,
\begin{equation}
    \ell(\mu_0, \mu_1, \Sigma, \pi_0, \pi_1) = \sum_{k=0}^{1} \sum_{(x, y) \in D_{t,k}} \bigg[ \log \pi_k -\frac{1}{2}\log |\Sigma| - \frac{1}{2} (x-\mu_k)^\top \Sigma^{-1}(x-\mu_k) \bigg] \notag + \constant
\end{equation}
where $D_{t, k}$ is the set of samples of $D_t$ that belongs to class $k$. Based on the likelihood function above, we can obtain the maximum likelihood estimates $\hat \mu_k, \hat \Sigma, \hat \pi_k$. The expression for the estimate $\hat \mu_k$ is
\begin{align}
    \hat \mu_k &= \frac{1}{|D_{t,k}|} \sum_{(x,y) \in D_{t, k}} x. \label{eq:mu_k}
\end{align}
Plugging these estimates into \cref{eqn:classifier}, we get,
\begin{align*}
    \hat{h}(x) &= \argmax_{k \in \{0, 1\}} \bigg[ \log \hat \pi_k -\frac{1}{2}\log |\hat \Sigma| -\frac{1}{2} (x-\hat \mu_k)^\top \hat \Sigma^{-1}(x-\hat \mu_k) \bigg] \\
    &= \argmax_{k \in \{0, 1\}} \bigg[ \log \hat \pi_k -\frac{1}{2}\log |\hat \Sigma| + x^\top \hat \Sigma^{-1} \hat \mu_k - \frac{1}{2} \hat \mu_k^\top \hat \Sigma^{-1} \mu_k \bigg]
\end{align*}
Therefore, $\hat{h}(x) = 1$ \text{iff}, 
\begin{align*}
    x^\top \hat \Sigma^{-1} \hat \mu_1 - \frac{1}{2} \hat \mu_1^\top \hat \Sigma^{-1} \mu_1 + \log \hat \pi_1 &> x^\top \hat \Sigma^{-1} \hat \mu_0 - \frac{1}{2} \hat \mu_0^\top \hat \Sigma^{-1} \mu_0 + \log \hat \pi_0 \\
    x^\top \hat \Sigma^{-1} \hat \mu_1 - x^\top \hat \Sigma^{-1} \hat \mu_0 &> \frac{1}{2} \hat \mu_1^\top \hat \Sigma^{-1} \mu_1 - \frac{1}{2} \hat \mu_0^\top \hat \Sigma^{-1} \mu_0 + \log \hat \pi_0 - \log \hat \pi_1 \\
    (\hat \Sigma^{-1}(\hat \mu_1-\hat \mu_0))^\top x &> (\hat \Sigma^{-1}(\hat \mu_1-\hat \mu_0))^\top \bigg( \frac{\hat \mu_0 + \hat \mu_1}{2} \bigg) + \log \frac{\hat \pi_0}{\hat \pi_1}
\end{align*}
Hence the FLD decision rule $\hat{h}(x)$ is 
\begin{equation}
    \hat h(x) = 
    \begin{cases}
        1, & \omega^\top x > c \\
        0, & \text{otherwise}
    \end{cases} \notag
\end{equation}
where $\omega = \hat \Sigma^{-1}(\hat \mu_1-\hat \mu_0)$ is a projection vector and $c = \omega^\top \big( \frac{\hat \mu_0 + \hat \mu_1}{2} \big) + \log \frac{\hat \pi_0}{\hat \pi_1}$ is a threshold. When $d = 1$ and $\pi_0 = \pi_1$, the decision rule reduces to
\begin{equation}\label{eqn:1d_fld_rule}
    \hat h(x) = 
    \begin{cases}
        1, & x >  \frac{\hat \mu_0 + \hat \mu_1}{2} \\
        0, & \text{otherwise}
    \end{cases}
\end{equation}

\subsection{Deriving the Generalization Error of the Target Distribution for Synthetic Data with FLD}\label{app:fld_err}

We would like to derive an expression for the average generalization error of the target distribution, when we consider the synthetic data described in~\cref{sec:app:synthetic}. For simplicity, we set the variance $\sigma^2$ of the class conditional densities of the synthetic data to $1$.

In the OOD-agnostic setting, the learning algorithm sees a single dataset $D = D_t \cup D_o$ of size $n+m$ which is a combination of both target and OOD samples. We can estimate $ \mu_k$ using~\cref{eq:mu_k} to obtain
\begin{equation}
    \aed{
    \hat \mu_k = \frac{1}{|D_k|} \sum_{(x,y) \in D_k} x &= \frac{\sum_{(x,y) \in D_{t, k}} x + \sum_{(x,y) \in D_{o, k}} x}{n_k + m_k}\\
    &= \frac{n_k \bar{x}_{t,k} + m_k \bar{x}_{o, k}}{n_k + m_k}\\
    &=  \frac{n \bar{x}_{t,k} + m \bar{x}_{o, k}}{n + m}.
    }
\end{equation}
where $D_k$ is the set of samples of $D$ that belongs to class $k$, $n_k = |D_{t, k}|$ and $m_k = |D_{o, k}|$ for $k \in \{ 0, 1 \}$. $\bar{x}_{t,k}$ and $\bar{x}_{o,k}$ denote the sample means of class $k$ in target and OOD datasets respectively. We assume that $\pi = \frac{1}{2}$ from which it follows that $n_k = n \pi_k = \frac{n}{2}$ and $m_k = m \pi_k = \frac{m}{2}$. We cannot explicitly compute $\bar{x}_{t,k}$ and $\bar{x}_{o,k}$ when the OOD samples are not explicitly known, because we cannot separate target samples from OOD samples in $D$.

Since the samples are drawn from Gaussians, their averages also follow Gaussian distributions. Hence, the threshold $\hat{c} = \frac{\hat \mu_0 + \hat \mu_1}{2}$ of the hypothesis $\hat h$, estimated using FLD, is a random variable with a Gaussian distribution i.e.,  $\hat{c} \sim \mathcal{N}(\mu_h, \sigma_h^2)$ where
\begin{align*}
    &\mu_h = \mathbb{E}[\hat{c}] = \frac{m \Delta}{n+m},\\
    &\sigma_h^2 = \Var[\hat{c}] = \frac{1}{n+m}.
\end{align*}
The target error of a hypothesis $\hat h$ is
\begin{align}
    p(\hat h(x) \neq y \mid x, \hat c) &= \frac{1}{2} p_{x \sim f_{t,1}}[x < \hat{c}] + \frac{1}{2} p_{x \sim f_{t,0}}[x > \hat{c}] \notag \\ 
    &= \frac{1}{2} + \frac{1}{2}p_{x \sim f_{t,1}}[x < \hat{c}] - \frac{1}{2}p_{x \sim f_{t,0}}[x < \hat{c}] \notag \\
    &= \frac{1}{2} \big[ 1 + \Phi ( \hat{c} - \mu ) - \Phi ( \hat{c} + \mu ) \big] \label{eq:single-head-risk}
\end{align}
Using~\cref{eq:single-head-risk}, the expected error on the target distribution $e_t(\hat h) = \mathbb{E}_{\hat{c} \sim \mathcal{N}(\mu_h, \sigma_h^2)}[ p(\hat h(x) \neq y \mid x, \hat c)]$ is given by,
\begin{align}
    e_t(\hat{h}) &= \int_{-\infty}^{\infty} \frac{1}{2} \big[ 1 + \Phi ( \hat{c} - \mu ) - \Phi ( \hat{c} + \mu ) \big] \frac{1}{\sigma_h} \phi \bigg( \frac{\hat{c} - \mu_h}{\sigma_h} \bigg) d\hat{c} \notag \\
    &= \int_{-\infty}^{\infty} \frac{1}{2} \big[ 1 + \Phi ( y\sigma_h + \mu_h - \mu ) - \Phi ( y\sigma_h + \mu_h + \mu ) \big] \phi(y) dy \notag \\
    &= \frac{1}{2}\bigg[ \Phi \bigg( \frac{\mu_h - \mu}{\sqrt{1 + \sigma_h^2}} \bigg)  + \Phi \bigg( \frac{-\mu_h - \mu}{\sqrt{1 + \sigma_h^2}} \bigg)\bigg] \notag
\end{align}
In the last equality, we make use of the identity $\int_{-\infty}^{\infty} \Phi(cx+d) \phi(x) dx = \Phi\big( \frac{d}{\sqrt{1 + c^2}} \big)$ where $\phi$ and $\Phi$ are the PDF and CDF of the standard normal. Substituting the expressions for $\mu_h, \sigma_h^2$ into the above equation, we get
\begin{equation}
    e_t(\hat{h}) = \frac{1}{2}\bigg[ \Phi \bigg( \frac{m \Delta - (n + m)\mu}{\sqrt{(n+m)(n+m+1)}} \bigg) + \Phi \bigg( \frac{-m \Delta - (n + m)\mu}{\sqrt{(n+m)(n+m+1)}} \bigg)\bigg]  \label{eq:single_head_identity}
\end{equation}
For synthetic data with $\sigma^2 \neq 1$, the target generalization error can be obtained by simply replacing $\mu$ and $\Delta$ with $\frac{\mu}{\sigma}$ and $\frac{\Delta}{\sigma}$ respectively in \cref{eq:single_head_identity}.

\subsection{OOD-Aware Weighted Fisher's Linear Discriminant}\label{app:weighted_fld}

We consider a target dataset $D_t = \{ (x_i, y_i) \}_{i=1}^n$ and an OOD dataset $D_o = \{ (x_i, y_i) \}_{i=1}^m$, which are samples from the synthetic data from \cref{sec:app:synthetic}. This setting differs from~\cref{app:fld_err} since we know whether each sample from $D = D_t \cup D_o$ is OOD or not. This difference allows us to consider a log-likelihood function that weights the target and OOD samples differently, i.e. we consider 
\begin{equation}
\scalemath{0.95}{
    \ell(\mu_0, \mu_1, \sigma_0^2, \sigma_1^2) = \sum_{k=0}^{1} \bigg( \alpha \sum_{(x, y) \in D_{t,k}} \bigg[ -\log \sigma_k - \frac{(x-\mu_k)^2}{2 \sigma_k^2} \bigg] + (1-\alpha) \sum_{(x, y) \in D_{o,k}} \bigg[ -\log \sigma_k  - \frac{(x-\mu_k)^2}{2 \sigma_k^2} \bigg]\bigg) + \constant.
    }
\end{equation}
$\alpha$ is a weight that controls the contribution of the OOD samples in the log-likelihood function. Under the above log-likelihood, the maximum likelihood estimate for $\mu_k$ is
\begin{equation}
    \hat \mu_k = \frac{\alpha \sum_{(x,y) \in D_{t, k}} x + (1 - \alpha) \sum_{(x,y) \in D_{o,k}} x}{\alpha|D_{t,k}| + (1 - \alpha)|D_{o,k}|}.
    \label{eqn:weighted_fld_estimates}
\end{equation}
We can make use of the above $\hat \mu_k$ to get a weighted FLD decision rule using~\cref{eqn:1d_fld_rule}.

\subsection{Deriving the Generalization Error of the Target Distribution for Synthetic Data with Weighted FLD}\label{app:weighted_fld_err}

We consider the synthetic distributions in~\cref{sec:app:synthetic} with $\sigma^2 = 1$. We re-write $\hat \mu_k$ from~\cref{eqn:weighted_fld_estimates} using notation from~\cref{app:fld_err}: 
\begin{equation}
    \hat \mu_k = \frac{n \alpha \bar{x}_{t,k} + m (1-\alpha)\bar{x}_{o, k}}{n \alpha + m (1-\alpha)} \notag.
\end{equation}
We can explicitly compute $\bar{x}_{t,k}$ and $\bar{x}_{o,k}$ in the OOD-aware setting since we can separate target samples from OOD samples. For the synthetic distribution, the threshold $\hat{c}_{\alpha} = \frac{\hat{\mu}_0 + \hat{\mu}_1}{2}$ of the hypothesis $\hat h_\alpha$ follows a normal distribution $\mathcal{N}(\mu_{h\alpha}, \sigma_{h\alpha}^2)$ where
\begin{align}
    &\mu_{h\alpha} = \mathbb{E}[\hat{c}_{\alpha}] = \frac{m(1-\alpha) \Delta}{n\alpha +m(1-\alpha)} \notag \\
    &\sigma_{h\alpha}^2 = \text{Var}[\hat{c}_{\alpha}] = \frac{\alpha^2 n + (1-\alpha)^2 m}{(\alpha n + (1-\alpha)m)^2} \notag
\end{align}
Similar to the~\cref{app:fld_err}, we derive an analytical expression for the expected target risk of the weighted FLD, which is
\begin{equation}
    e_t(\hat{h}_{\alpha}) = \frac{1}{2}\bigg[ \Phi \bigg( \frac{\mu_{h\alpha} - \mu}{\sqrt{1 + \sigma_{h\alpha}^2}} \bigg)  + \Phi \bigg( \frac{-\mu_{h\alpha} - \mu}{\sqrt{1 + \sigma_{h\alpha}^2}} \bigg)\bigg]\label{eqn:weighted_fld_gen_err}
\end{equation}

\subsection{Additional Experiments using FLD}
\label{app:fld_simualtions}

\begin{figure}[htbp]
  \centering
  \newcommand{\widtha}{0.25}
  \newcommand{\widthb}{0.26}
  \includegraphics[width=\textwidth]{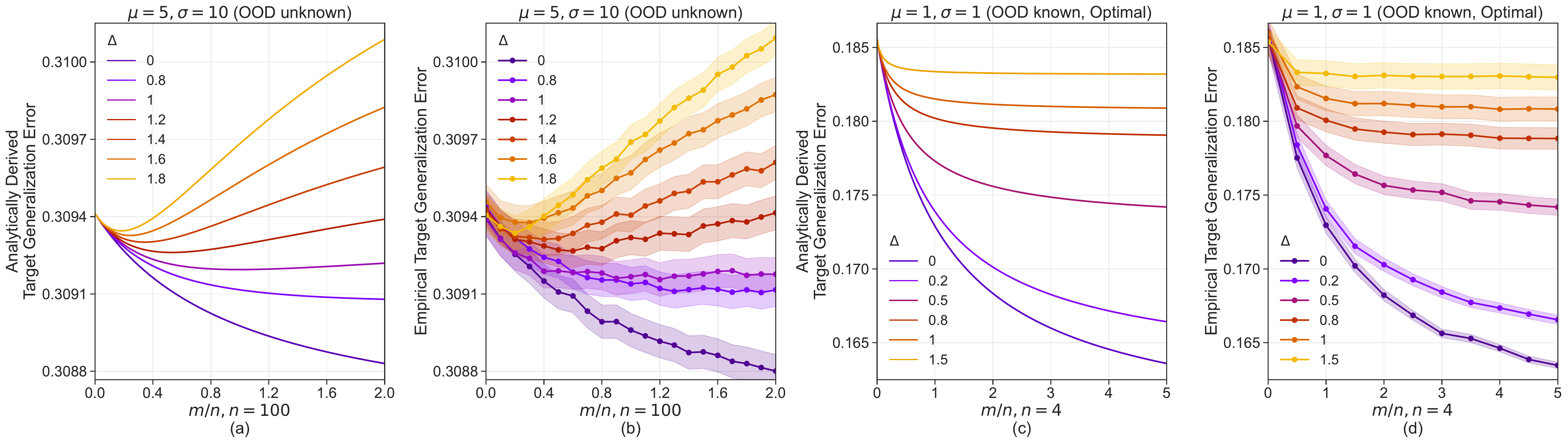}
  \caption{The FLD generalization error (Y-axis) on the target distribution is plotted against the ratio of OOD samples to target samples (X-axis). Figures (a) and (c) are plotted using the analytical expressions in \cref{eq:single_head_identity} and \cref{eqn:weighted_fld_gen_err} respectively while figures (b) and (d) are the corresponding plots from Monte-carlo simulations. The Monte-carlo simulations agree with the plots from the analytical expression, which validates its correctness. \textbf{(a) and (b):} The figure is identical to~\cref{fig:gauss_tasks} and considers synthetic data with $n=100$, $\mu=5$ and $\sigma=10$ in the OOD-agnostic setting. While a small number of OOD samples improves generalization on the target distribution, lots of samples increase the generalization error on the target distribution. \textbf{(c) and (d):} The figures consider synthetic data with  $n=4$, $\mu=1$ and $\sigma=1$ in the OOD-aware setting. If we consider the weighted FLD trained with optimal $\alpha^*$, then the average generalization error monotonically decreases with more OOD samples. Shaded regions indicate 95\% confidence intervals over the Monte-Carlo replicates.}
\end{figure}

\subsection{Deriving the Upper Bound in~\cref{thm:ben_david} for the OOD-Agnostic Fisher's Linear Discriminant}
\label{s:app:fld_upper_bound}

We begin by defining the following quantities: Given a hypothesis $h : X \rightarrow \{0, 1\}$, the probability according to the distribution $P_s$ that $h$ disagrees with a labeling function $f$ is defined as,
\[
e_s(h, f) = \mathbb{E}_{x \sim P_s}\left[ | h(x) - f(x) | \right]
\]
For a hypothesis space $H$, \citep{ben2010theory} defines the divergence measure between two distributions $P_t$ and $P_o$ in the symmetric difference hypothesis space as,
\[
d_H^\ast (P_t, P_s) = 2 \sup_{h, h' \in H} | e_s(h, h') - e_t(h, h') |
\]
With these definitions in place, we restate a slightly modified version of the Theorem 3 from~\citep{ben2010theory} below.
\pagebreak
\begin{theorem}
Let $H$ be a hypothesis space of VC dimension $d$. Let $D$ be a dataset generated by drawing $n$ samples from a target distribution $P_t$ and $m$ OOD samples from $P_o$. If $\hat h \in H$ is the empirical minimizer of $\alpha e_{t}(h) + (1-\alpha) e_{o}(h)$ on $D$ and $h^\ast_t = \min_{h \in H} e_t(h)$ is the target error minimizer, then for any $\delta \in (0, 1)$, with probability as least $1-\delta$ (over the choice of samples), 
\begin{equation}
    e_{t}(\hat h) \leq \underbrace{e_t(h^\ast_t) + 4 \sqrt{\frac{\alpha^2}{n} + \frac{(1-\alpha)^2}{m}} \sqrt{2 d \log (2(n+m+1)) + 2 \log(\frac{8}{\delta})} + 2 (1-\alpha) \bigg( \frac{1}{2} d_H (P_t, P_o) + \lambda \bigg)}_{U(n,m, d_H (P_t, P_o))} 
    \label{eq:original_bound}
\end{equation}
where, $\lambda$ is the combined error of the ideal joint hypothesis given by $h^\ast = \argmin_{h \in H} e_t(h) + e_s(h)$. Hence, $\lambda = e_t(h^\ast) + e_s(h^\ast)$.
\end{theorem}
We wish to adapt the above theorem according to our FLD example in~\cref{s:fld_example} and consequently find an expression for the upper bound $U(n,m, d_H (P_t, P_o))$ in terms of $n,m$ and $\Delta$. As we do not know of the existence of OOD samples in dataset $D$, we find the hypothesis $\hat h$ by minimizing the empirical loss below.
\begin{align*}
    \hat e(h) &= \frac{1}{n + m} \sum_{i=1}^{n+m} \ell(h(x_i), y_i) \\
    &= \frac{1}{n+m} \sum_{(x,y) \in D_t} \ell(h(x), y) + \frac{1}{n+m} \sum_{(x,y) \in D_o} \ell(h(x), y) \\
    &= \frac{n}{n+m} e_t(h) + \frac{m}{n+m} e_o(h).
\end{align*}
Here, we have assumed that $\ell(\cdot)$ is the 0-1 loss. Therefore, under the OOD agnostic setting, we minimize the objective function $e(h) = \alpha e_{t}(h) + (1-\alpha) e_{o}(h)$ where $\alpha = n/(n+m)$. Since we deal with a univariate FLD, the VC dimension of the hypothesis space is equal to $d = 1+ 1 = 2$. Plugging these terms in \cref{eq:original_bound}, we can rewrite the upper bound as,
\begin{equation}
    U(n,m, d_H (P_t, P_o)) = e_t(h^\ast_t) + 4 \sqrt{4 \log (2(n+m+1)) + 2 \log(\frac{8}{\delta})} + \frac{2m}{n+m} \bigg( \frac{1}{2} d_H (P_t, P_o) + \lambda \bigg)
    \label{eq:intermediate_bound}
\end{equation}
The first term of the above expression corresponds to the error of the best hypothesis $h^\ast_t$ in class $H$ for the target distribution $P_t$. Thus, $e_t(h^\ast_t)$ is equivalent to the Bayes optimal error or the lowest possible error achievable for the target distribution, under $H$. By setting $m = 0$ in \cref{eq:single_head_identity}, we arrive at the expected error $e_t(\hat h)$ on the target distribution when we estimate $\hat h$ using $n$ target samples. The Bayes optimal error $e_t(h^\ast_t)$ is then equal to the limit of $e_t(\hat h)$ as $n \rightarrow \infty$. 
\[
 e_t(h^\ast_t) = \lim_{n \rightarrow \infty} e_t(\hat h) = \lim_{n \rightarrow \infty} \Phi \bigg( - \frac{n(\mu/\sigma)}{\sqrt{n(n+1)}} \bigg) = \Phi (-\mu / \sigma)
\]
Intuitively, the threshold corresponding to the ideal joint hypothesis $h^\ast$ for our FLD example is given by the mid point between the centers of the two distributions,
\[
h^\ast(x) = \argmin_{h \in H} e_o(h) + e_t(h) = \mathbbm{1}_{(\Delta/2, \infty)}(x)
\]
where $I_A(x)$ is the indicator function of the subset $A$.
Therefore, the combined error $\lambda$ of the ideal joint hypothesis can be computed as follows.
\begin{align*}
    \lambda &= e_o(h^\ast) + e_t(h^\ast) \\
    &= \frac{1}{2} p_{x \sim f_{t, 0}} [x > \Delta / 2] + \frac{1}{2} p_{x \sim f_{t, 1}} [x < \Delta / 2] + \frac{1}{2} p_{x \sim f_{o, 0}} [x > \Delta / 2] + \frac{1}{2} p_{x \sim f_{o, 1}} [x < \Delta / 2] \\
    &= \Phi \bigg( \frac{- \Delta / 2 - \mu}{\sigma} \bigg) + \Phi \bigg( \frac{\Delta / 2 - \mu}{\sigma} \bigg)
\end{align*}
Finally, we turn to the divergence term $d_H(P_t, P_o)$. Let $h, h' \in H$ be two hypotheses with thresholds $c$ and $c'$, respectively. From the definition of $e_t(h, h')$ we have,
\begin{align*}
    e_t(h, h') &= \mathbb{E}_t\left[ |h(x) - h'(x)| \right] \\
    &= \mathbb{E}_t\left[ |\mathbbm{1}_{(c, \infty)}(x) - \mathbbm{1}_{(c', \infty)}(x)| \right] \\
    &= \mathbb{E}_t\left[ \mathbbm{1}_{(\min(c, c'), \max(c, c')]}(x) \right] \\
    &= p_t [\min(c, c') < x \leq \max(c, c')]  \\
    &= \frac{1}{2} p [ x \leq \max(c, c') \mid y = 0] + \frac{1}{2} p [ x \leq \max(c, c') \mid y = 1] - \frac{1}{2} p [ x \leq \min(c, c') \mid y = 0] - \frac{1}{2} p [ x \leq \min(c, c') \mid y = 1]  \\
    &= \frac{1}{2} \bigg[ \Phi \bigg( \frac{\max(c, c') + \mu}{\sigma} \bigg) + \Phi \bigg( \frac{\max(c, c') - \mu}{\sigma} \bigg) - \Phi \bigg( \frac{\min(c, c') + \mu}{\sigma} \bigg) - \Phi \bigg( \frac{\min(c, c') - \mu}{\sigma} \bigg) \bigg] \\
    &= \psi_{\mu, \sigma}(c, c')
\end{align*}
Similarly, we can show that $e_o(h, h') = \psi_{\mu, \sigma}(c-\Delta, c'-\Delta)$. Therefore, we can rewrite the expression for $d_H(P_t, P_o)$ as follows.
\[
d_H(P_t, P_o) =  2 \sup_{h, h' \in H} | e_o(h, h') - e_t(h, h') | = 2 \sup_{c, c' \in [0, \Delta]} |  \psi_{\mu, \sigma}(c-\Delta, c'-\Delta) - \psi_{\mu, \sigma}(c, c') | = d_H^\ast (\Delta)
\]
Using this expression we can numerically compute $d_H^\ast$, given the values of $\mu, \sigma$ and $\Delta$. Plugging in the expressions we have obtained for $ e_t(h^\ast_t), \lambda$ and $d_H(P_t, P_o)$ in \cref{eq:intermediate_bound}, we arrive at the desired upper bound for the expected target error $e_t(\hat h)$ of our FLD example.
\begin{equation}
     U(n,m,\Delta) = \Phi (-\mu / \sigma) + 4 \sqrt{4 \log (2(n+m+1)) + 2 \log(\frac{8}{\delta})} + \frac{2m}{n+m} \bigg[ \frac{1}{2}  d_H^\ast (\Delta) +  \Phi \bigg( \frac{- \Delta / 2 - \mu}{\sigma} \bigg) + \Phi \bigg( \frac{\Delta / 2 - \mu}{\sigma} \bigg) \bigg] 
     \label{eq:final_bound}
\end{equation}

\subsection{Comparisons between the Upper Bound and the True Target Generalization Error}
\label{s:app:additional_bound_results}

\begin{figure}[h]
    \centering
    \includegraphics[width=\linewidth]{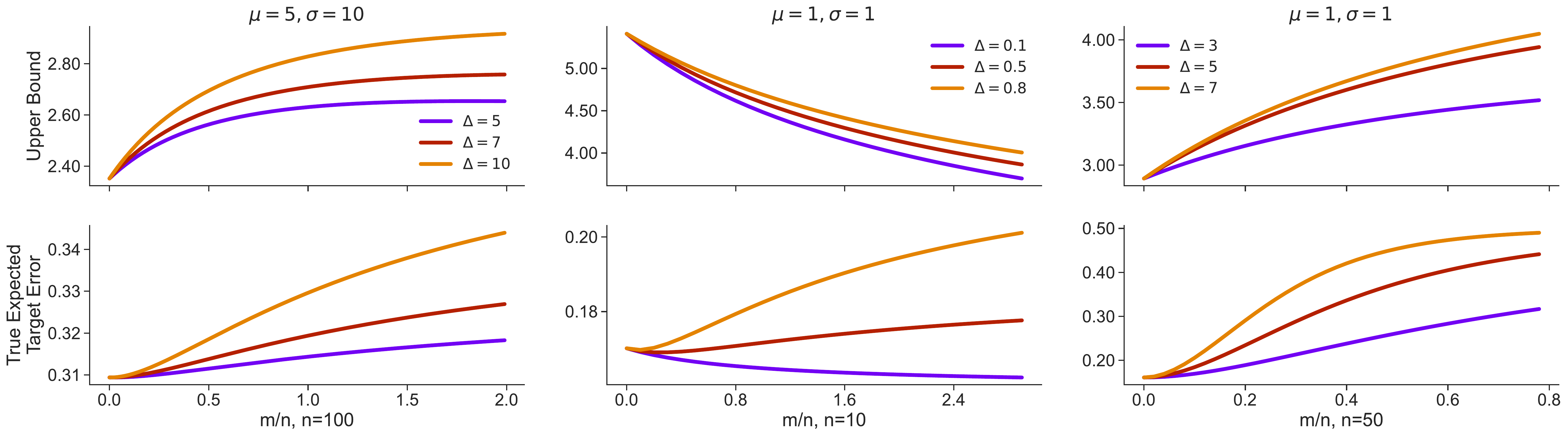}
    \caption{The upper bound (as computed by \cref{eq:final_bound}) and the true expected target error (as computed by \cref{eq:single_head_identity}), for 3 different variations of the FLD example in \cref{s:fld_example}. In the left and right columns, we observe that the shape of the curve agrees somewhat with that of the true error. Notice that the separation $\Delta$ between the distributions of the target and OOD data is large in these cases. \cref{fig:bound1} and the middle column of the current figure indicate that the upper bound does not exhibit a non-monotonic trend while the true error does. It is also important to note that the bound is significantly vacuous in all cases. These observations suggest that the Theorem 3 from the work of~\citet{ben2010theory} does not explain the non-monotonic trends that we have identified in this work.}
    \label{fig:additional_bound_results}
\end{figure}

\section{Experiments with Neural Networks}
\label{sec:app:neural_net}

\subsection{Datasets}
\label{sec:app:image}

We experiment on images from CIFAR-10, CINIC-10~\citep{darlow2018cinic} and several datasets from the DomainBed benchmark~\citep{gulrajani2020search}: Rotated MNIST~\citep{ghifary2015domain}, PACS~\citep{li2017deeper}, and DomainNet~\citep{peng2019moment}. We construct sub-tasks from these datasets as explained below.

\paragraph{CIFAR-10} We use of tasks from Split-CIFAR10~\citep{zenke_continual_2017} which are five binary classification sub-tasks constructed by grouping consecutive labels of CIFAR-10. The 5 task distributions are airplane vs. automobile ($T_1$), bird vs. cat ($T_2$), deer vs. dog ($T_3$), frog vs. horse ($T_4$) and ship vs truck ($T_5$). All the images are of size $(3,32,32)$.

\paragraph{CINIC-10} This dataset combines CIFAR-10 with downsampled images from ImageNet. It contains images of size $(3,32,32)$ across $10$ classes (same classes as CIFAR-10). As there are two sources of the images within this dataset, it is a natural candidate for studying distribution shift. The construction of the dataset motivates us to consider two distributions from CINIC-10: \begin{enumerate*}[{(1)}] \item Distribution with only CIFAR images, and \item Distribution with only ImageNet images \end{enumerate*}.

\paragraph{Rotated MNIST} This dataset is constructed from MNIST by rotating the images (which are of size $(1,28,28)$. All MNIST images rotated by an angle  $\theta^\circ$ are considered to belong to the same distribution. Hence, we can consider the family of distributions which is characterized by 10-way classification of hand-written digit images rotated $\theta^\circ$. By varying $\theta$, we can obtain a number of different distributions.

\paragraph{PACS} PACS contains images of size $(3,224,244)$ with $7$ classes present across 4 domains $\{$art, cartoons, photos, sketches$\}$. In our experiments, we consider only $3$ classes ($\{$Dog, Elephant, Horse$\}$) out of the $7$ and consider the 3-way classification of images from a given domain as a distribution. Therefore, we can have a total of $4$ distinct distributions from PACS.

\paragraph{DomainNet} Similar to PACS, this dataset contains images of size $(3,224,244)$ from 6 domains $\{$ clipart, infograph, painting, quickdraw, real, sketches$\}$ across $345$ classes. In our experiments, we consider only $2$ classes, ($\{$Bird, Plane$\}$) and consider the binary classification of images from a given domain as a distribution. As a result, we can have a total of $6$ distinct distributions from PACS.

\subsection{Forming Target and OOD Distributions}
\label{app:form-tar-ood}

We consider two types of setups to study the impact of OOD data:

\paragraph{OOD data arising due to geometric intra-class nuisances} We study the effect of intra-class nuisances using a classification task using samples from a target distribution and OOD samples from a transformed version of the same distribution. In this regard, we consider the following experimental setups.

\begin{enumerate}
    \item \textbf{Rotated MNIST: unrotated images as target and $\theta^\circ$- rotated images as OOD:} We consider the 10-way classification (see~\cref{sec:app:image}) of unrotated images as the target data and that of the $\theta^\circ$- rotated images as the OOD data. We can have different OOD data by selecting different values for $\theta$.
    
    \item \textbf{Rotated CIFAR-10: $\bm{T_2}$ as target and rotated $\bm{T_2}$ as OOD:} We choose the bird vs. cat ($T_2$) task from Split-CIFAR10 as the target distribution. We then rotate the images of $T_2$ by an angle $\theta^\circ$ counter-clockwise around their centers to form a new task distribution denoted by $\theta$-$T_2$, which we consider as OOD. Different OOD datasets can be obtained by selecting different values for $\theta$.

    \item \textbf{Blurred CIFAR-10: $\bm{T_4}$ as target and blurred $\bm{T_4}$ as OOD:} We choose the Frog vs. Horse ($T_4$) task from Split-CIFAR10 as the target distribution. We then add Gaussian blur with standard deviation $\sigma$ to the images of $T_4$ to form a new task distribution denoted by $\sigma$-$T_2$, which we consider as the OOD. By setting distinct values for $\sigma$, we have different OOD datasets.
\end{enumerate}

\paragraph{OOD data arising due to category shifts and concept drifts}  We study this aspect using two different target and OOD classification problems as described below.

\begin{enumerate}
    \item \textbf{Split-CIFAR10: $\bm{T_i}$ as Target and $\bm{T_j}$ as OOD:} We choose a pair of distinct tasks from the 5 binary classification tasks of Split-CIFAR10 and consider one as the target distribution and the other as the OOD. We perform experiments for all pairs of distributions (20 in total) in Split-CIFAR10.
    \item \textbf{PACS: Photo-domain as target and X-domain as OOD:} Out of the four 3-way classification tasks from PACS described in~\cref{sec:app:image}, we select the photo-domain as the target distribution and consider one of the remaining 3 domains (for instance, the sketch-domain) as the OOD.
    \item \textbf{DomainNet: Real-domain as target and X-domain as OOD:} Out of the six binary classification tasks from DomainNet described in~\cref{sec:app:image}, we consider the real-domain as the target distribution and select one of the remaining 5 domains (for instance, the painting-domain) as OOD.
    \item \textbf{CINIC-10: CIFAR10 as target and ImageNet as OOD:} Here we simply select the 10-way classification of CIFAR images as the target distribution and that of ImageNet as OOD.
\end{enumerate}

\subsection{Experimental Details}\label{app:expt_details}

In the above experiments, for each random seed, we randomly select a fixed sample of size $n$ from the target distribution. Next, we select OOD samples of varying sizes $m$ such that the previous samples are a subset of the next set of samples. The samples from both target and OOD distributions preserve the ratio of the classes. For rotated MNIST, rotated CIFAR-10, and blurred CIFAR-10, when selecting multiple sets of OOD samples, the OOD images that correspond to the $n$ selected target images are disregarded. For PACS and DomainNet, the images are downsampled to $(3,64,64)$ during training.

For both the OOD-agnostic (OOD unknown) and OOD-aware (OOD known) settings, at each $m$-value, we construct a combined dataset containing the $n$ sized target set and $m$ sized OOD set. We use a CNN (see~\cref{sec:app:arch}) for experiments in the both of these settings. We experiment with $\alpha$ fixed to $0.5$ (naive OOD-aware model) and with the optimal $\alpha^\ast$. We average the runs over 10 random seeds and evaluate on a test set comprised of only target samples.

In the optimal OOD-aware setting, we use a grid-search to find the optimal $\alpha^\ast$ for each value of $m$. We use an adaptive equally-spaced $\alpha$ search set of size $10$ such that it ranges from $\alpha^\ast_{prev}$ to $1.0$ (excluding $1.0$) where $\alpha^\ast_{prev}$ is the optimal value of $\alpha$ corresponding to the previous value of $m$. We use this search space since we expect $\alpha^*$ to be an increasing function of $m$. 

\subsection{Neural Architectures and Training}\label{sec:app:arch}
We primarily use 3 different network architectures in our experiments: (a) a small convolutional network with 0.12M parameters (denoted by \emph{SmallConv}), (b) a wide residual network~\citep{zagoruyko2016wide} of depth 10 and widening factor 2 (WRN-10-2), and (c) a larger wide residual network of depth 16 and widening factor 4 (WRN-16-4). SmallConv comprises of 3 convolution layers (kernel size 3 and 80 filters) interleaved with max-pooling, ReLU, batch-norm layers, with a fully-connected classifier layer in our experiments. 

\begin{figure}[h]
\centering
\includegraphics[width=0.8\linewidth]{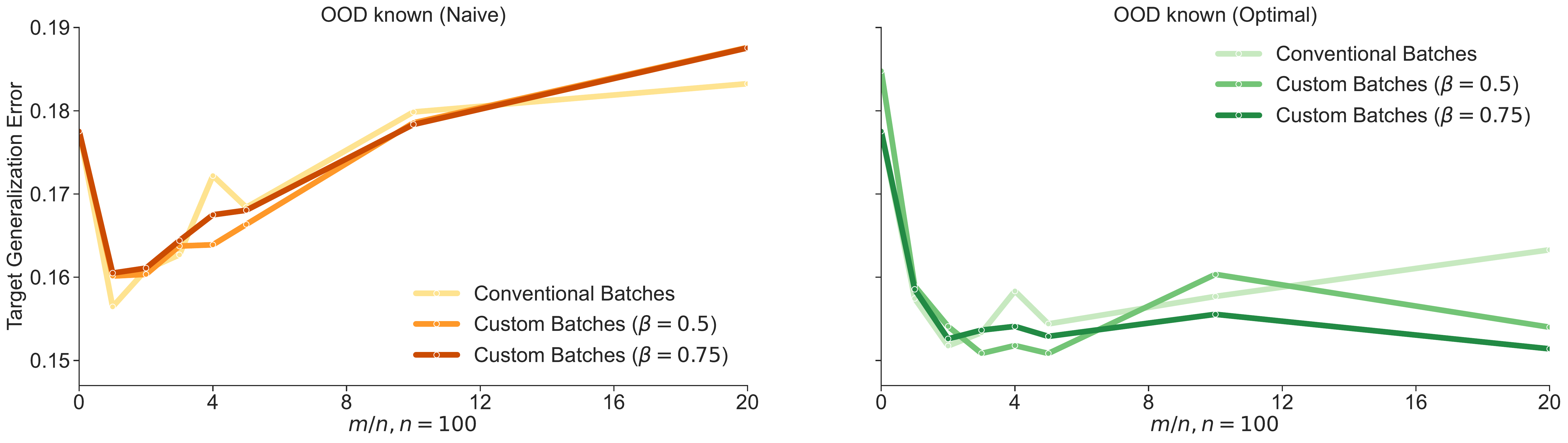}
\caption{\textbf{Standard mini-batching strategy versus ensuring that every mini-batch has a fraction $\beta$ samples from the target distribution.} The test error of a neural network (SmallConv) on the target distribution (Y-axis) is plotted against the number of OOD samples (X-axis) for the target-OOD pair of $T_1$ and $T_5$. One set of curves (lightest shade of green and yellow) considers mini-batches which are constructed using sampling without replacement; This is the standard strategy used in supervised learning. The other curves consider $\beta = 0.5$ (intermediate shades of orange and green) and $\beta = 0.75$ (darkest shade of red and green). All plots are in the OOD-aware setting. \textbf{Left:} If we consider $\alpha=0.5$, then the choice of $\beta$ has little effect on the generalization error. \textbf{Right:} However, if we use $\alpha^*$ to weight the OOD and target losses, then the generalization error depends on the the choice of $\beta$ with $\beta=0.75$ having the lowest test error.}
\label{fig:beta_comparison}
\end{figure}

\cref{tab:exp_summary} provides a summary of network architectures used in the experiments described earlier. All the networks are trained using stochastic gradient descent (SGD) with Nesterov's momentum and cosine-annealed learning rate. The hyperparameters used for the training are, learning rate of 0.01, and a weight-decay of $10^{-5}$. All the images are normalized to have mean 0.5 and standard deviation 0.25. In the OOD-agnostic setting, we use sampling without replacement to construct the mini-batches. In the OOD-aware settings (both naive and optimal), we construct mini-batches with a fixed ratio of target and OOD samples. See \cref{sec:app:minibatch} and \cref{fig:beta_comparison} for more details.

\begin{table}[!htbp]
    \vspace*{1em}
    \centering
    \footnotesize
    \renewcommand{\arraystretch}{1.25}
    \begin{tabular}{llrrrr}
        \toprule
        \textbf{Experiment} & \textbf{Network(s)} & \textbf{\# classes} & \textbf{n} & \textbf{ Image Size} & \textbf{Mini-Batch Size} \\
        \midrule
        Rotated MNIST       &  SmallConv           & 10  & 100 & (1,28,28) & 128 \\
        Rotated CIFAR-10    &  SmallConv, WRN-10-2 & 2 & 100 & (3,32,32) & 128 \\
        Blurred CIFAR-10    &  WRN-10-2            & 2 & 100 & (3,32,32) & 128 \\
        Split-CIFAR10       &  SmallConv, WRN-10-2 & 2 & 100 & (3,32,32) & 128 \\
        PACS                &  WRN-16-4            & 3 & 30  & (3,64,64) & 16 \\
        DomainNet           &  WRN-16-4            & 2 & 50  & (3,64,64) & 16 \\
        CINIC-10            &  WRN-10-2            & 10 & 100 & (3,32,32) & 128 \\
        \bottomrule         
    \end{tabular}
    \caption{Summary of network architectures used in the experiments}
    \label{tab:exp_summary}
\end{table}

\subsection{Construction of Mini-Batches}\label{sec:app:minibatch}

Consider a mini-batch $\{ (x_{b_i}, y_{b_i}) \}_{i=1}^{B}$ of size $B$. Let the randomly chosen mini-batch contains $B_t$ target samples and $B_o$ OOD samples ($B = B_t + B_o$). Let $\hat{e}_{B,t}(h)$ and $\hat{e}_{B,o}(h)$ denote the average mini-batch surrogate losses for the $B_t$ target samples and $B_o$ OOD samples respectively.

In the OOD-aware (when we know which samples are OOD) setting, $\hat{e}_{B,t}(h)$ and $\hat{e}_{B,o}(h)$ can be computed explicitly for each mini-batch resulting in the mini-batch gradient 
\begin{equation}
    \hat \nabla  \hat{e}_B(h) =
    \alpha \hat \nabla \hat{e}_{B,t}(h) + (1 - \alpha) \hat \nabla \hat{e}_{B,o}(h).
\end{equation}
If we were to sample without replacement, we expect the fraction of the target samples in every mini-batch to approximately equal $\frac{n}{n+m}$ on average. However, if $m >> n$, we run into a couple of issues. First, we observe that most mini-batches have no target samples, making it impossible to compute $\hat \nabla \hat e_{B, t}(h)$. Next, even if the mini-batch does have some target samples, there are very few of them, resulting in high variance in the estimate $\hat \nabla \hat e_{B, t}(h)$. 

Hence, we find it beneficial to consider alternative sampling schemes for the mini-batch. Independent of the values of $n$ and $m$, we use a sampler which ensures that every mini-batch has a fixed fraction of target samples, which we denote by $\beta$. For example if the mini-batch size $B$ is $20$ and if  $\beta=0.5$, then every mini-batch has $10$ target samples and $10$ OOD samples regardless of $n$ and $m$. Note that this sampling biases the gradient, but results in reduced variance estimates. In practice, we observe improved test errors when we set $\beta$ to either $0.5$ or $0.75$.

\subsection{Additional Experiments with Neural Networks}
\label{app:splitcifar10_task_matrix}

\begin{figure}[H]
\centering
\includegraphics[width=0.5\linewidth]{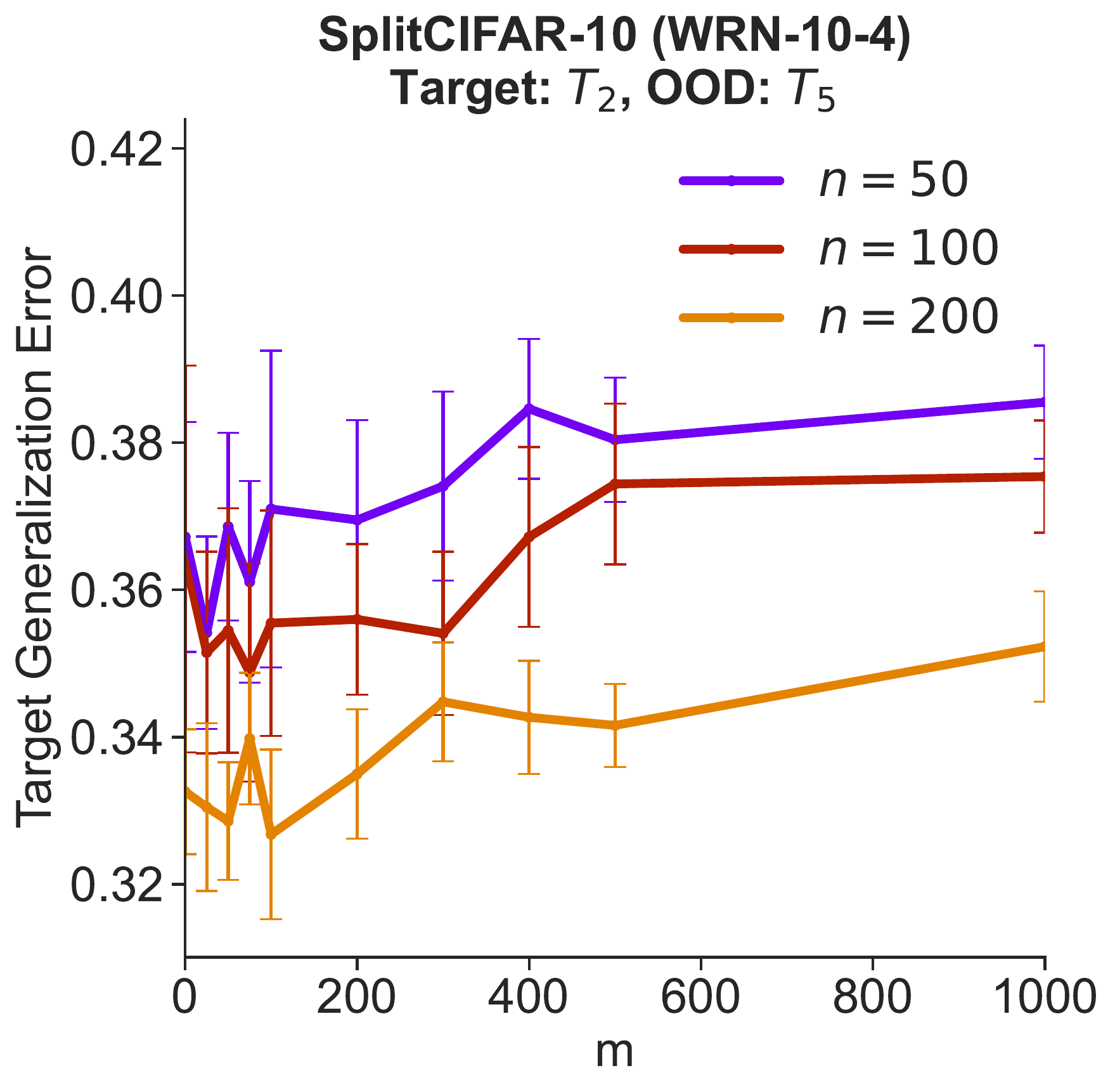}
\caption{We plot the generalization error on the target distribution (Y-axis) against the number of OOD samples $m$ (X-axis) across three different target sample sizes, $n = 50, 100$ and $200$ for the target-OOD pair $T_2$ and $T_5$ from Split-CIFAR10. Non-monotonic trends in generalization error are present in all the three cases. The trend is less apparent for $n=50$ since the number of samples is small resulting in a large variance. Error  bars indicate 95\% confidence intervals (10 runs).}
\label{fig:multiple_n_exp}
\end{figure}

\begin{figure}[H]
\centering
\subfloat[\centering]{\includegraphics[width=0.78\linewidth]{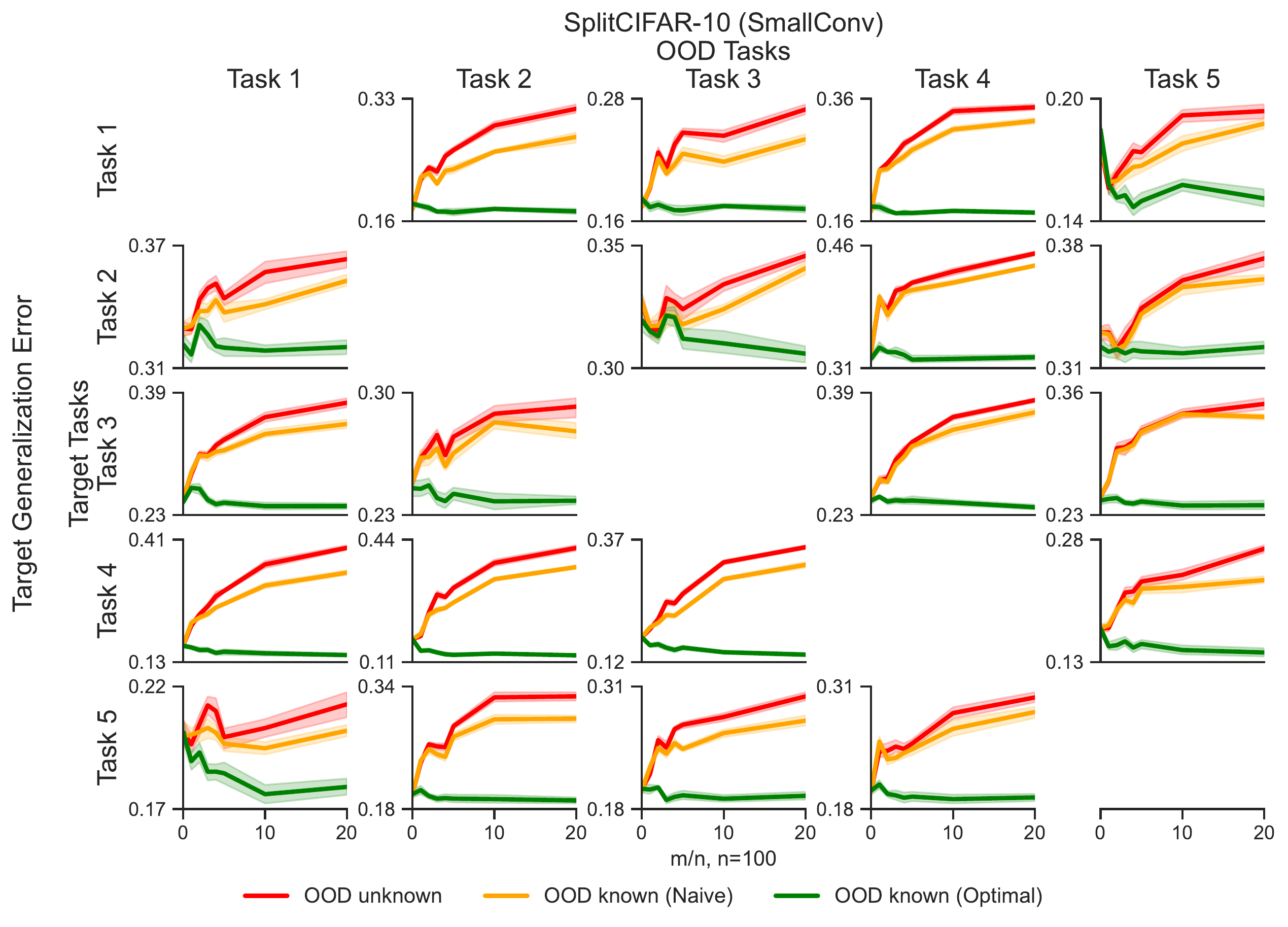}} \\
\subfloat[\centering]{\includegraphics[width=0.78\linewidth]{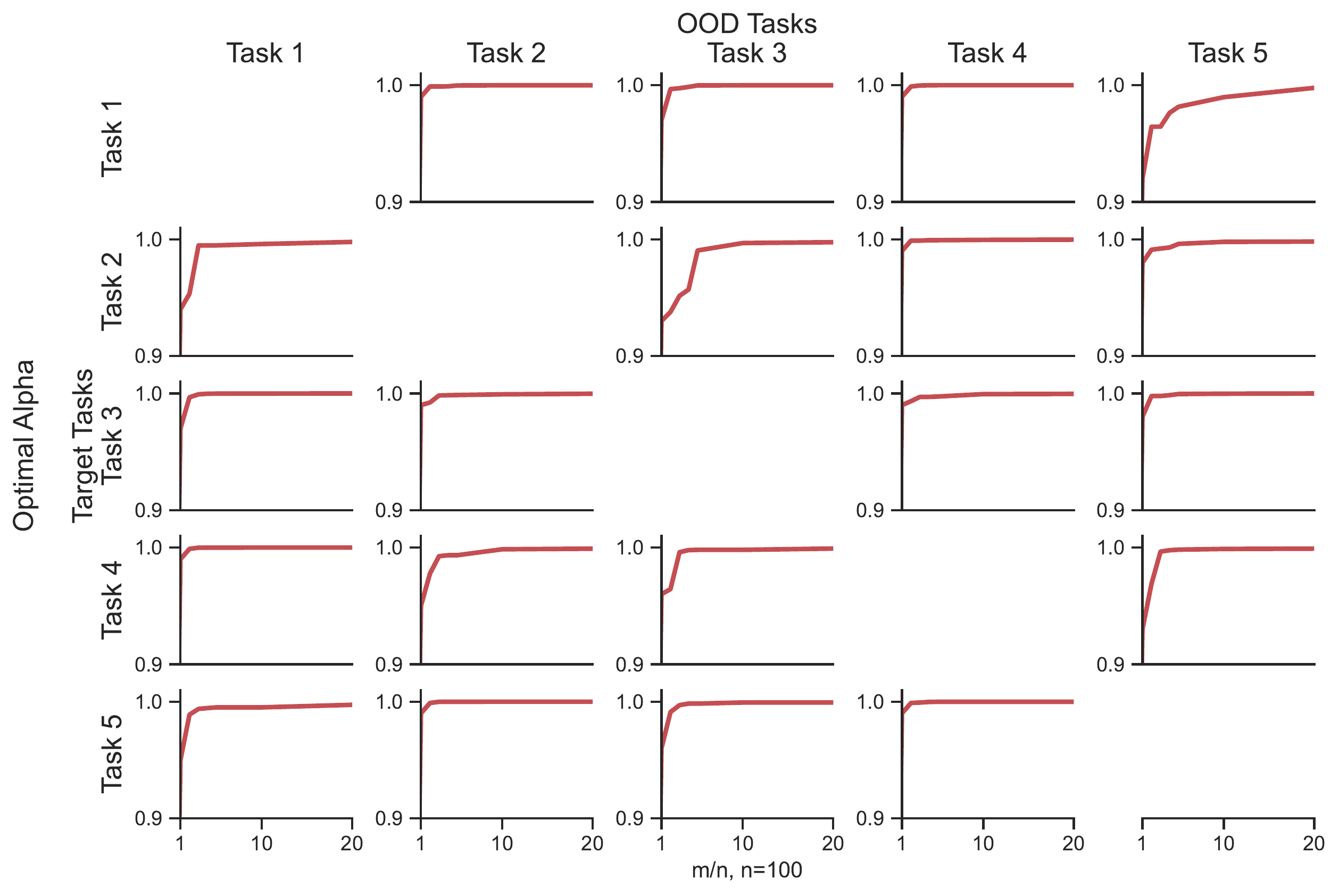}}
\caption{\textbf{(a)} We plot the test error of SmallConv on the target distribution (Y-axis) against the ratio of number of OOD samples to the number of samples from the target task (X-axis), for all target-OOD pairs from Split-CIFAR10. A neural net trained with a loss weighted by $\alpha^*$ is able to leverage OOD data to improve the networks ability to generalize on the target distribution.  Shaded regions indicate 95\% confidence intervals over 10 experiments. \textbf{(b)} The optimal $\alpha^*$ (Y-axis) is plotted against the number of OOD samples (X-axis) for the optimally weighted OOD-aware setting. As we increase the number of OOD samples, we see that $\alpha^*$ increases. This allows us to balance the variance from few target samples and the bias from using OOD samples from a different disitribution.}
\label{fig:cifar1_task_matrix}
\end{figure}

\begin{figure}[H]
\centering
\subfloat[\centering]{\includegraphics[width=0.78\linewidth]{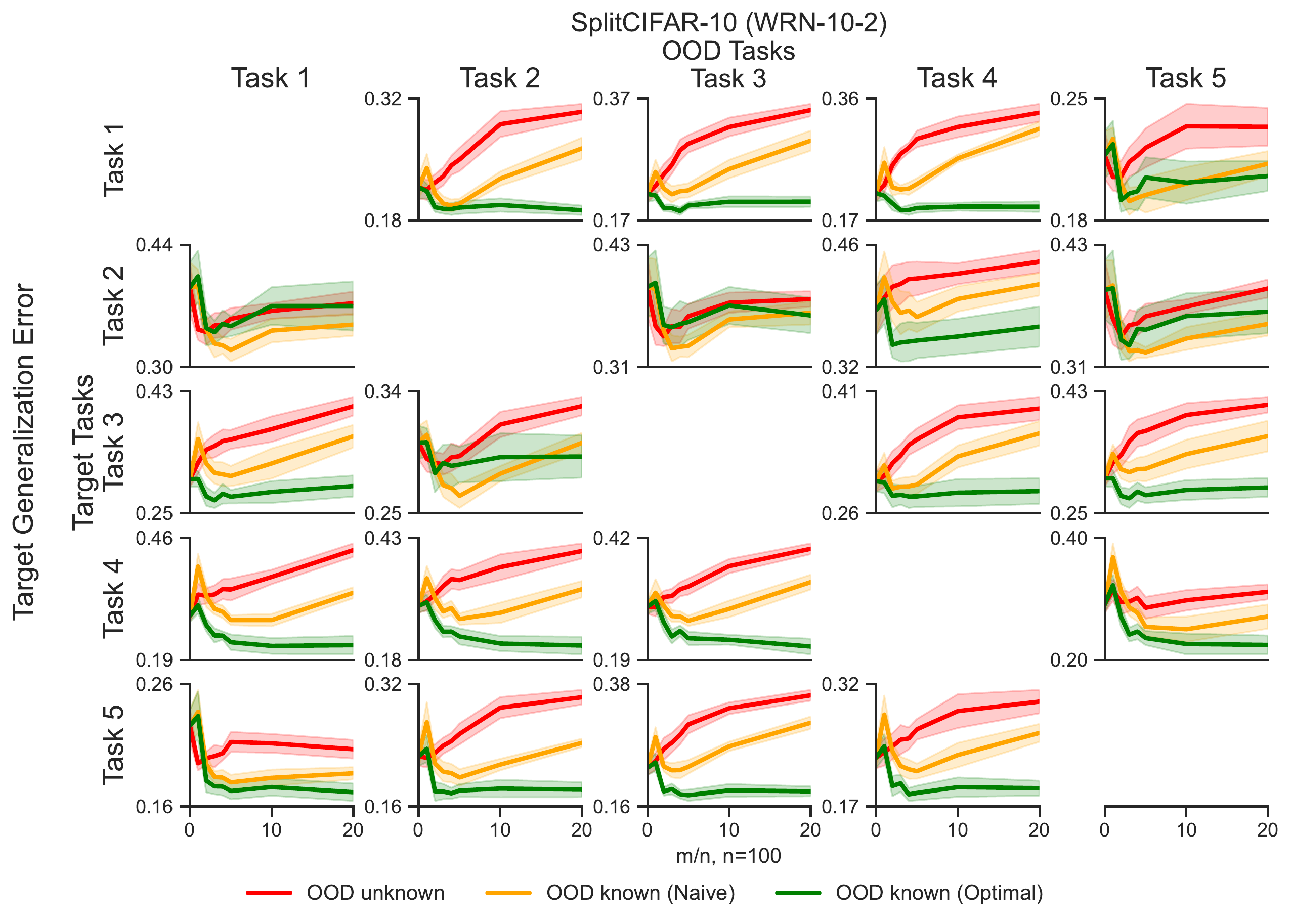}} \\
\subfloat[\centering]{\includegraphics[width=0.78\linewidth]{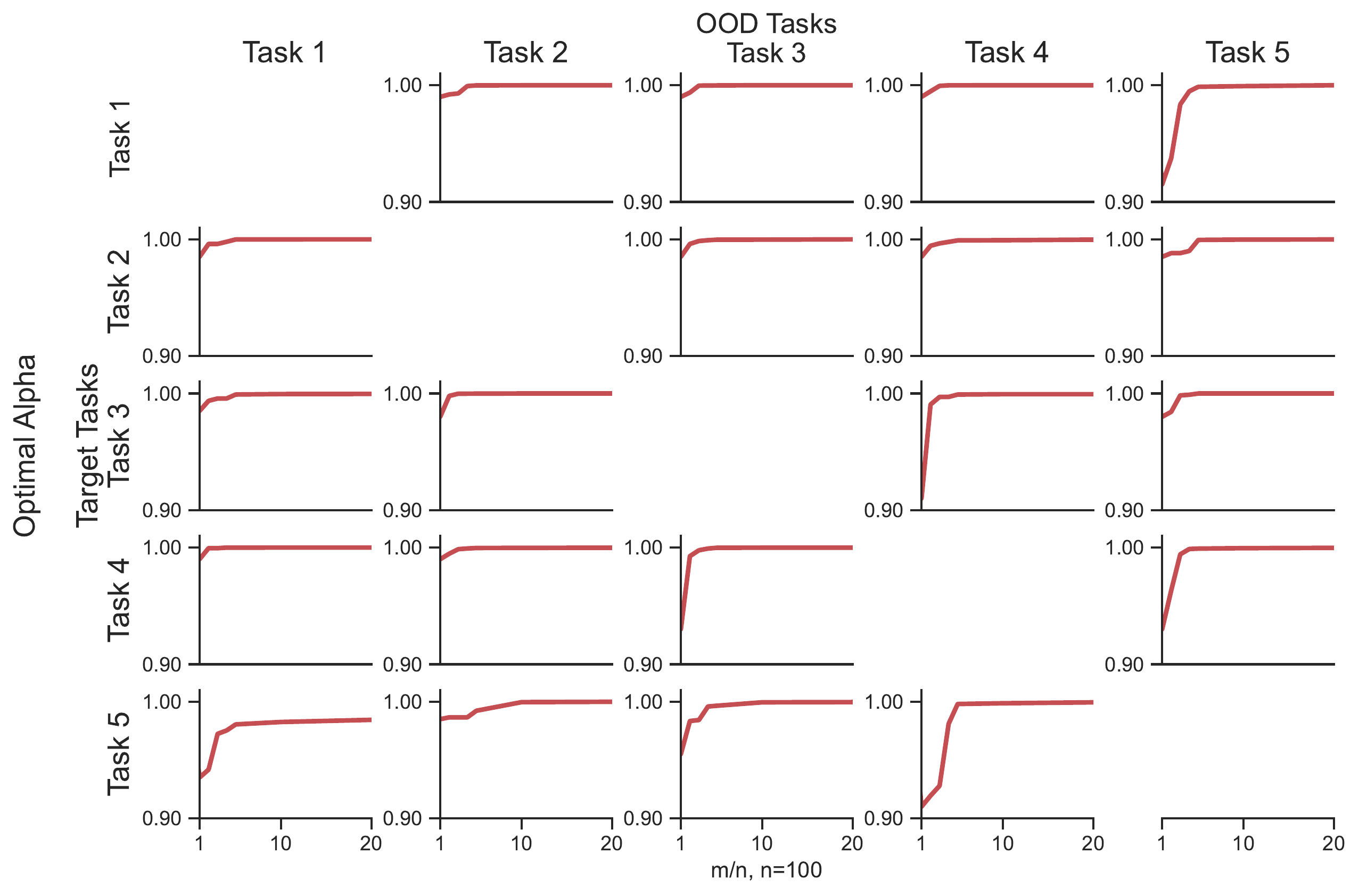}}
\caption{\textbf{(a)} We plot the test error of WRN-10-2 on the target distribution (Y-axis) against the ratio of number of OOD samples to the number of samples on the target task (X-axis), for all target-OOD pairs from Split-CIFAR10. A neural net trained with a loss weighted by $\alpha^*$ is able to leverage OOD data to improve the networks ability to generalize on the target distribution.  Shaded regions indicate 95\% confidence intervals over 10 experiments. \textbf{(b)} The optimal $\alpha^*$ (Y-axis) is plotted against the number of OOD samples (X-axis) for the optimally weighted OOD-aware setting. As we increase the number of OOD samples, we see that $\alpha^*$ increases. This allows us to balance the variance from few target samples and the bias from using OOD samples from a different disitribution.}
\label{fig:cifar2_task_matrix}
\end{figure}

\begin{figure}[!t]
\centering
\includegraphics[width=\linewidth]{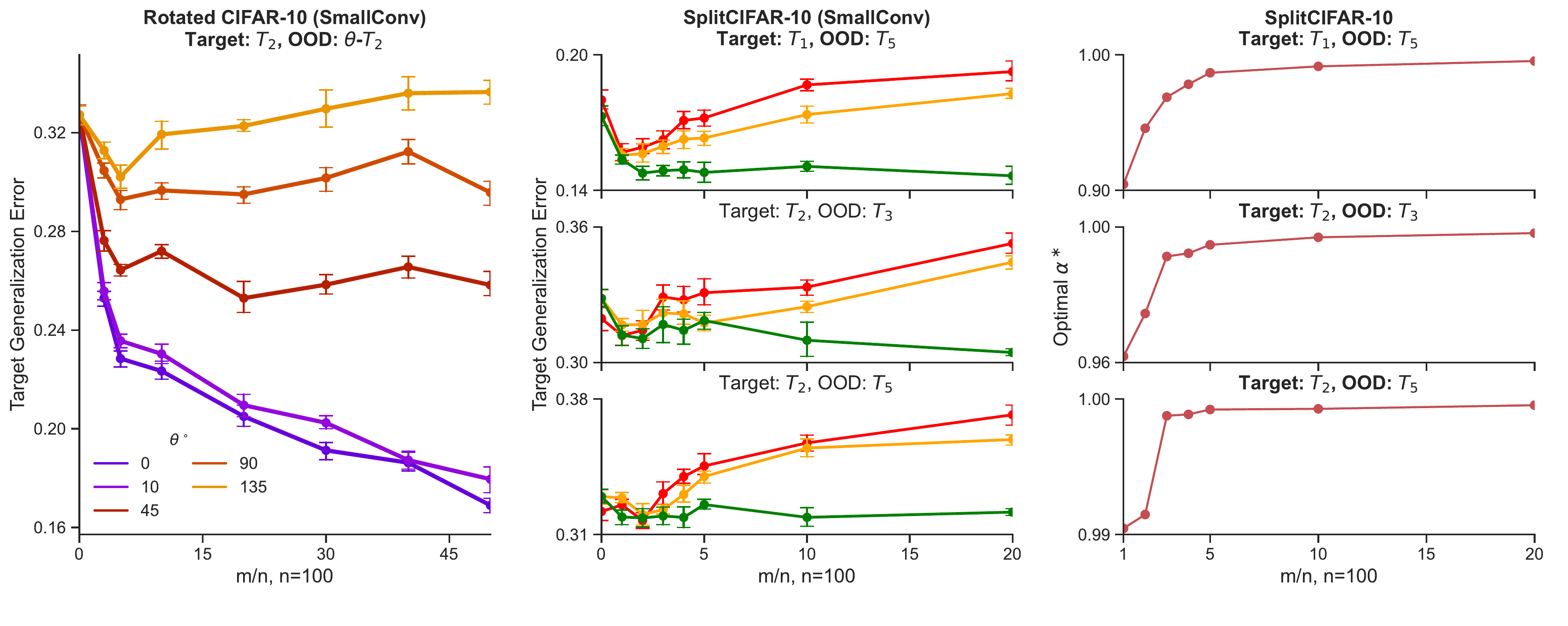}
\caption{\textbf{Left:} A binary classification problem (Bird vs. Cat) is the target distribution and images of these classes rotated by different angles $\theta^\circ$ are OOD. We see non-monotonic curves for larger values of $\theta^\circ$. For $135^\circ$ in particular, the generalization error at $m/n=50$ is worse than the generalization error with no OOD samples, i.e. OOD samples actively hurt generalization.\\[0.25em]
\textbf{Middle:} Generalization error on the target distribution is plotted against the number of OOD samples for 3 different target-OOD pairs constructed from CIFAR-10 for three settings: OOD-agnostic ERM where we minimize the total average risk over both distributions (red), an objective which minimizes the sum of the average loss of the target and OOD distributions which corresponds to $\a=1/2$ (OOD-aware, yellow) and an objective which minimizes an optimally weighted convex combination of the target and OOD empirical loss (green).\\[0.25em]
\textbf{Right:} The optimal $\a^\ast$ obtained via grid search for the three problems in the middle column plotted against different number of OOD samples. Note that the appropriate value of $\a$ lies very close to 1 but it is never exactly 1. In other words the OOD samples always benefit if we use the weighted objective in~\cref{thm:ben_david}, even if this benefit is marginal in cases when OOD samples are very different from those of the target.
}
\label{fig:selected_rotated_and_dual_tasks}
\end{figure}

\begin{figure}[H]
\centering
\includegraphics[width=0.8\linewidth]{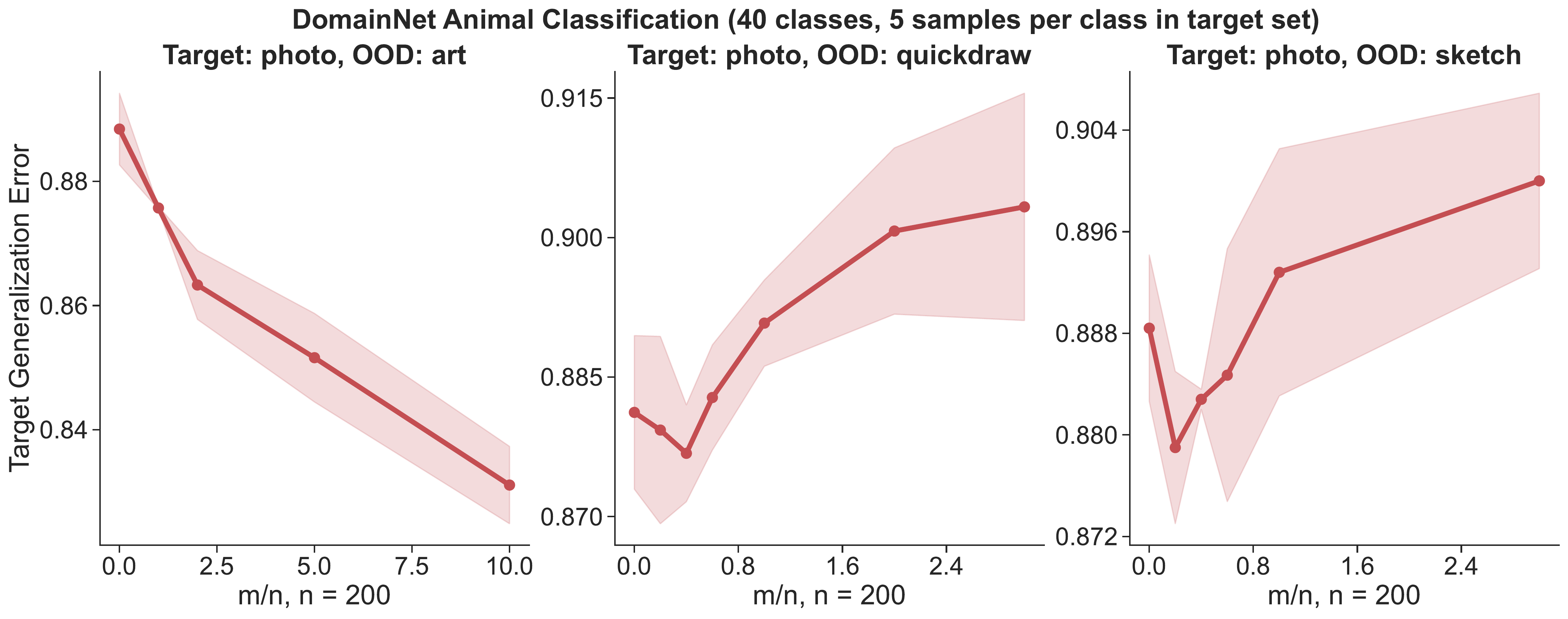}
\caption{We consider a 40-class classification problem from DomainNet where the classes are animals from three super-classes: mammals, cold blooded animals and birds. The target distribution considers images of animals from the "real" domain. OOD data considers images from the domains ``paintings'', ``quickdraw'' and ``sketches''. We plot the target generalization error against the ratio of OOD and target samples and observe the risk to be non-monotonic for 2 of the 3 OOD domains. Note that the error of the trained network (0.85) is lower than the error of a classifier that predicts all classes with uniform probability (0.975). The error is high because we use very few training samples; the number of target samples is 200 (i.e. only 5 samples per class). Note that the error bars indicate 95\% confidence intervals over 3 runs.}
\label{fig:domainnet-40}
\end{figure}

\end{document}